\documentclass[sigconf,screen]{acmart}
\usepackage{subcaption}
\usepackage{tabularx}
\usepackage{enumitem}
\usepackage{balance}

\usepackage{xcolor}
\newcommand{\rev}[1]{#1}

\setlist{nosep, leftmargin=*} 

\AtBeginDocument{%
  }

\setcopyright{cc}
\setcctype{by}
\acmDOI{10.1145/3757279.3788663}
\acmYear{2026}
\copyrightyear{2026}
\acmISBN{979-8-4007-2128-1/2026/03}
\acmConference[HRI '26]{Proceedings of the 21st ACM/IEEE International Conference on Human-Robot Interaction}{March 16--19, 2026}{Edinburgh, Scotland, UK}
\acmBooktitle{Proceedings of the 21st ACM/IEEE International Conference on Human-Robot Interaction (HRI '26), March 16--19, 2026, Edinburgh, Scotland, UK}
\received{2025-09-30}
\received[accepted]{2025-12-23}

\begin{document}


\title[POIROT: Investigating Direct Tangible vs. Digitally Mediated Interaction and Attitude Moderation...]{POIROT: Investigating Direct Tangible vs. Digitally Mediated Interaction and Attitude Moderation in Multi-party Murder Mystery Games}

\author{Wen Chen}
\orcid{0009-0002-4204-8777}
\affiliation{%
  \institution{Tsinghua University}
  \city{Beijing}
  \country{China}
}
\email{chenwen_bot@mail.tsinghua.edu.cn}

\author{Rongxi Chen}
\orcid{0009-0009-7298-7223}
\affiliation{%
  \institution{Tsinghua University}
  \city{Beijing}
  \country{China}
}
\email{vichenxi@vanguardadvisories.com}

\author{Shankai Chen}
\orcid{0009-0002-3237-0383}
\affiliation{%
  \institution{University of Pennsylvania}
  \city{Philadelphia}
  \country{USA}
}
\email{skchen@seas.upenn.edu}

\author{Huiyang Gong}
\orcid{0009-0000-5211-0525}
\affiliation{%
  \institution{Syracuse University}
  \city{Syracuse}
  \country{USA}
}
\email{hgong11@syr.edu}

\author{Minghui Guo}
\orcid{0009-0004-9298-8438}
\affiliation{%
  \institution{National University of Singapore}
  \city{Singapore}
  \country{Singapore}
}
\email{mguo77@u.nus.edu}

\author{Yingri Xu}
\orcid{0009-0008-1495-5058}
\affiliation{%
  \institution{Royal College of Art}
  \city{London}
  \country{United Kingdom}
}
\email{yingri.xu@network.rca.ac.uk}

\author{Xintong Wu}
\orcid{0009-0009-6397-034X}
\affiliation{%
  \institution{Tsinghua University}
  \city{Beijing}
  \country{China}
}
\email{wuxt22@mails.tsinghua.edu.cn}

\author{Xinyi Fu}
\authornote{Corresponding Author.}
\orcid{0000-0001-6927-0111}
\affiliation{%
  \institution{Tsinghua University}
  \city{Beijing}
  \country{China}
}
\email{fuxy@tsinghua.edu.cn}

\renewcommand{\shortauthors}{Chen et al.}

\begin{abstract}
As social robots take on increasingly complex roles like game masters (GMs) in multi-party games, the expectation that physicality universally enhances user experience remains debated. This study challenges the "one-size-fits-all" view of tangible interaction by identifying a critical boundary condition: users' Negative Attitudes towards Robots (NARS). In a between-subjects experiment (\emph{N} = 67), a custom-built robot GM facilitated a multi-party murder mystery game (MMG) by delivering clues either through direct tangible interaction or a digitally mediated interface. Baseline multivariate analysis (MANOVA) showed no significant main effect of delivery modality, confirming that tangibility alone does not guarantee superior engagement. However, primary analysis using multilevel linear models (MLM) revealed a reliable moderation: participants high in NARS experienced markedly lower narrative immersion under tangible delivery, whereas those with low NARS scores showed no such decrement. Qualitative findings further illuminate this divergence: tangibility provides novelty and engagement for some but imposes excessive proxemic friction for anxious users, for whom the digital interface acts as a protective social buffer. These results advance a conditional model of HRI and emphasize the necessity for adaptive systems that can tailor interaction modalities to user predispositions.
\end{abstract}

\ccsdesc[500]{Human-centered computing~Interaction design theory, concepts and paradigms}
\ccsdesc[500]{Human-centered computing~Empirical studies in collaborative and social computing}
\ccsdesc[500]{Human-centered computing~Empirical studies in HCI}

\keywords{Human-Robot Interaction, Social Robots, Multi-Party Interaction, Tangible Interaction, Social Deduction Games, Negative Attitudes towards Robots (NARS)}

\maketitle

\section{Introduction}
The role of social robots is expanding rapidly beyond utilitarian functions and into the realm of complex human social life, involving education, entertainment, and collaborative play \cite{belpaeme2018social, munoz2021robo, rato2023robots}. One compelling application is their use as a game master (GM) in traditional tabletop and narrative-driven games \cite{fischbach2018follow, kennington2014probabilistic}. Building upon this, we turn to a more complex and underexplored frontier: the domain of murder mystery games (MMGs) \cite{zhu2024player, wu2023deciphering}. This genre, relying heavily on deception, reasoning, logical deduction, and intricate social interaction, has arisen as a challenging testbed for studying group Human-Robot Interaction (HRI) \cite{moujahid2022multi, parreira2022design, tennent2019micbot, oliveira2021human}. Although recent studies in large language models (LLMs) have fostered interest in virtual GMs \cite{sasha2025multi, song2024you}, the unique challenges of designing effective interaction modalities for a robot in such a socially and cognitively demanding game have received limited empirical attention \cite{munoz2021robo, rato2023robots, lin2022benefits, ng2024role}. To this end, we introduce Plot-Oriented Interactive Robot for Organized Theatrics (POIROT)\footnote{The acronym POIROT alludes to Agatha Christie's fictional detective Hercule Poirot, reflecting the robot's role as a narrative orchestrator in murder mystery settings.}, a custom-built robot designed to facilitate this multi-party social deduction game.

Designing effective interaction modalities for such a system in complex settings remains an open challenge \cite{tsai2025assistive, azevedo2024comfort, 9613601}. Although POIROT is a physically embodied robot, simply considering whether a robot is present or not is insufficient for understanding how users engage with it \cite{tsai2025assistive, sanfilippo2025caged, lagomarsino2023maximising, takayama2009influences}. Recent shifts in HRI emphasize that how an interaction is delivered, including the degree of immediacy, proximity, and mediation, is often more consequential than embodiment alone \cite{tsai2025assistive, jung2025proximity, Su2023RecentAI}. For instance, Sanfilippo et al. argue that contemporary transitions from basic HRI to collaboration and teaming prioritize the design of interaction modalities like shared space, physical contact, and multi-modal signaling. In multi-party contexts, a central design decision concerns the choice between direct tangible engagement (e.g., the robot moving toward players and handing over physical items) and digitally mediated interaction (e.g., distributing information via tablet interfaces) \cite{nigro2025social, han2024co, tan2022group}. These modalities generate distinct social dynamics: direct physical interaction elevates proximity and immediacy, but also imposes stronger proxemic demands, which can potentially exacerbate social pressure for users sensitive to interaction at close range \cite{tsai2025assistive, mead2015proxemics, mumm2011human, rajamohan2019factors}. In contrast, mediated interaction provides a buffered and lower-pressure channel that reduces the perceived intensity of the robot's presence and offers greater predictability \cite{pandey2025integrating, gittens2024remote, han2024co}. Consequently, the key research question turns from whether embodiment matters to how these contrasting delivery modalities, direct vs. mediated, shape users' experiences in complex group settings \cite{kabacinska2025influence, esterwood2025virtually, gillet2021robot, han2024co}.

These differences in modality are of particular impact for users who vary in their comfort with robots \cite{kabacinska2025influence, herzog2025influence, li2019comparing}. Negative Attitudes towards Robots Scale (NARS), which captures stable predispositions such as anxiety, avoidance tendencies, and discomfort with unpredictable robot behaviors, closely underpins how individuals manage proxemic boundaries and perceived social demands \cite{nomura2006measurement, nomura2007experimental, kodani2024effects, kabacinska2025influence}. Under a psychological distance framework, users with high NARS may find direct tangible interaction as intrusive, effortful, or difficult to predict, leading to elevated proxemic stress and a reduced sense of control \cite{liberman2007psychological, trope2007construal, herzog2025influence, miller2021more, nomura2007experimental}. Conversely, digitally mediated delivery can function as a psychological buffer via increasing perceived distance, lowering social pressure, and enabling more comfortable engagement \cite{shahid2024effectiveness, osler2023sociality, warnock2020treating}. Prior work has shown that user attitudes shape dyadic HRI, however, much less is known about how these predispositions interact with interaction modality in co-located, multi-party environments, where social coordination demands are substantially higher \cite{kabacinska2025influence, nigro2025social, dahiya2023survey, schneiders2022non}. This study therefore examines how interaction modality and individual predispositions jointly influence user experience in a richly constructed narrative game setting.

To investigate these dynamics, we conducted a between-subjects experiment ($N=67$) in which groups of players participated in a multi-party MMG facilitated by POIROT. We examined two theoretically grounded interaction modalities: direct physical delivery and digitally mediated delivery, which were separately operationalized as the Physical-Card condition (POIROT approached players and dispensed printed clue cards) and the In-App condition (the same clues presented via a custom tablet interface). By holding the robot’s identity, dialogue, narrative pacing, and overall game flow constant, this design enables a clean comparison of modality effects while simultaneously minimizing confounding differences in behavior or content. The specific hypotheses derived from the theoretical framework are presented after the related work in Section 2.4.

Through this approach, the contributions of our study are threefold. Empirically, we present a novel controlled evaluation of interaction modality in a complex, co-located multi-party MMG facilitated by a robot, comparing direct physical delivery with digitally mediated delivery under tightly standardized conditions. At the theoretical level, we provide a psychological distance account that helps explain how proxemic demands and user attitudes in conjunction shape experience. From a design perspective, our findings highlight the limits of “one-size-fits-all” approaches and underscore the need for adaptive interaction modalities that adjust proximity and mediation based on user predispositions.

\vspace{-8pt}
\section{Related Work}
\subsection{Robots as Game Masters and Facilitators}
The application of social robots in gaming and entertainment is a growing area of HRI, \rev{moving beyond task assistance and toward richer social roles} \cite{rato2023robots, munoz2021robo, sandoval2022human}. Previous research \rev{has demonstrated the potential of robots to improve children's personalized learning and therapy as partners or guides \cite{ligthart2023design, shiomi2021two, kouroupa2022use, rakhymbayeva2021long}. In cooperative or adversarial games, robots have been designed as teammates that facilitate collaboration and moderate conflict, or as opponents capable of sophisticated strategy \cite{correia2018group, nichols2021collaborative, lombardi2025would}. Informed by these positions in structured games,} researchers have devoted increasing attention to robots in narrative and facilitation roles, such as mediators or characters in storytelling and puzzle-solving games, in which robots have been demonstrated to enhance immersion and enjoyment \cite{lin2022benefits, ng2024role}. 

In addition to diegetic roles, robots have also operated as non-diegetic facilitators of group dynamics through regulating turn-taking, promoting fairness, and orchestrating participation by means of social cues such as gaze, proxemics, and movement \cite{short2017robot, short2015towards, mutlu2009footing, buchem2024human}. While these studies substantiate the promise of robots in managing multi-party interaction, they mostly focus on isolated facilitation mechanics rather than the intricate demands defining social deduction games \cite{lin2022benefits, ng2024role, short2017robot}.

With the emergence of murder mystery games (MMGs) as a promising but challenging research frontier \cite{bowman2025transformative, wu2023deciphering, zhu2024player, nonomura2025speaks}, many efforts have primarily adopted LLM-based disembodied agents to simulate reasoning and narrative control in such games \cite{wu2023deciphering, zhu2024player, lombardi2025would, nonomura2025speaks}. However, very few studies have examined how a physical robot might fulfill the full suite of a GM's responsibilities encompassing arbitration of rules, handling of physical game artifacts, and management of private information \cite{lin2022benefits, ng2024role, kennington2014probabilistic}. \rev{This gap is striking given that the GM's authority in MMGs is not only cognitive but also material—managing the flow of tangible clues and information \cite{bowman2025transformative, xiong2020murder}. In this work, we expand on these parallel threads and address this critical and underexplored niche by investigating how a robot GM's approach to delivering game materials, whether through direct tangible actions or digitally mediated alternatives, impacts player experience in a complex multi-party MMG.}

\vspace{-5pt}
\subsection{Direct vs. Mediated Interaction}
In multi-party HRI environments, the effects of a robot's behavior are shaped by both what the robot communicates and how the content is delivered \cite{10.1145/3643803, bazzano2018human, rodriguez2015bellboy, li2015benefit}. Recent work has highlighted that the modality of interaction such as the level of immediacy, proximity, and mediation, is instrumental in shaping users' comfort, attention, and overall experience, which transcends earlier binary contrasts between physical and virtual agents and centers on the concrete social and psychological demands embedded in dissimilar forms of delivery \cite{leichtmann2022personal, tsai2025assistive, gittens2024remote, zu2024let}.

Direct tangible interaction, as exemplified by a robot physically approaching a user and presenting an object, inherently increases spatial immediacy \cite{rahmah2025modeling, hall1966hidden, leichtmann2022personal, tsai2025assistive}. Drawing from Hall’s proxemics theory, such close-range actions can heighten attentional salience and psychological intensity, while also imposing greater proxemic demands that may induce social pressure or discomfort for some users \cite{hall1966hidden, rodriguez2015bellboy}. In these circumstances, the robot's movement and physical closeness are not merely functional behaviors but also social signals requiring users to manage attention, expectations, and personal space \cite{li2015benefit, rodriguez2015bellboy}.

In comparison, digitally mediated delivery introduces a layer of psychological distance between the user and the robot \cite{xu2018investigating, gittens2024remote, walther1996computer}. Theories of Computer-Mediated Communication (CMC) describe mediated channels as reducing immediacy and filtering out potentially stressful cues through a form of social buffering \cite{walther1996computer, walther2011theories}. Emerging HRI literature similarly shows that mediated interfaces can mitigate perceived social pressure and enhance predictability, allowing users to engage with the robot's content without the demands resulting from direct physical interaction \cite{gittens2024remote, xu2018investigating}.

These perspectives indicate that immediate and mediated channels constitute fundamentally different interaction demands \cite{li2015benefit, walther1996computer}. Instead of assuming greater immediacy always improves engagement, it is crucial to understand how users with varying sensitivity react to the proxemic overload of direct interaction in contrast with the buffered structure of mediated interaction \cite{tsai2025assistive, zojaji2025impact, fitter2020closeness, mumm2011human}.

\vspace{-6pt}
\subsection{Attitudes toward Robots and Sensitivity to Proxemic Demands}
User comfort with robots exhibits considerable variation, a trait well-captured by the Negative Attitudes towards Robots Scale (NARS), which measures anxiety and avoidance regarding unpredictable behaviors \cite{nomura2006measurement, ivaldi2015towards, naneva2020systematic}. Theoretically, high NARS users prioritize reduced immediacy and greater predictability to manage interaction-based intrusions \cite{syrdal2009negative, nomura2006measurement, niewrzol2024development}. Consequently, direct tangible actions, such as entering personal space to physically pass an object, impose elevated proxemic demands that these users find socially taxing \cite{rahmah2025modeling, nenna2024addressing, miller2021more, obaid2016stop}. For example, Rahmah et al. show that tasks like object handovers elicit markedly higher proxemic sensitivity in anxious individuals, necessitating greater separation distance for sense of ease \cite{rahmah2025modeling}. By contrast, digitally mediated delivery offers a protective buffer, reducing real-time coordination and aligning with preferences for lower immediacy \cite{gittens2024remote, seo2025evaluating, van2022impact}. While NARS is widely studied in dyadic settings, its specific interaction with concrete delivery modalities in complex, co-located multi-party environments where social coordination demands are notably higher is still underexplored \cite{nigro2025social, muller2024egocentric, amirova202110}.

\vspace{-8pt}
\subsection{The Present Study and Hypotheses}
Guided by the above theoretical perspectives on interaction modality and psychological distance, the present study investigates how the method of delivery (direct tangible vs. digitally mediated) shapes user experience in a co-located multi-party setting, and how this relationship is conditioned by individual predispositions toward robots.

Drawing from traditional presence-centered interpretations, physical delivery may enhance immediacy and thus increase engagement and immersion. To test this baseline assumption in our specific context, we first examine the main effect of interaction modality:

\textbf{H1 (Baseline Effect of Modality):} The Physical-Card condition (representing direct tangible delivery) will lead to greater narrative immersion, game experience, and social presence compared to the In-App condition (representing digitally mediated delivery).

However, our primary theorized mechanism concerns the moderating role of individual attitudes. Based on the psychological distance account, we argue that the social buffering introduced by digital mediation will be advantageous for users predisposed to discomfort in high-immediacy or close-range situations. Therefore, we hypothesize:

\textbf{H2 (Moderating Role of NARS):} Users' NARS will moderate the effects of the delivery modality. Specifically, for users with high NARS, the Physical-Card condition will result in lower narrative immersion and game experience compared to the In-App condition due to increased proxemic demands, whereas this difference will be attenuated or negligible for users with low NARS.

Collectively, these hypotheses enable us to test both a presence-oriented assumption (H1) and the central prediction derived from the psychological distance account (H2).

\vspace{-8pt}
\section{Methods}
\subsection{Participants}
A total of 72 participants were recruited from a Tsinghua University community through campus posters and posts on social media channels. Five participants who did not complete all the study questionnaires were excluded from the analysis. This resulted in a final sample of 67 participants (40 female, 26 male, 1 preferred not to disclose). The majority of participants were young adults; 6.0\% ($n=4$) were aged 18--20, 70.1\% ($n=47$) were aged 21--30, 20.9\% ($n=14$) were aged 31--40, and 3.0\% ($n=2$) were over 40. Most of the participants were local residents (95.5\%, $n=64$), with three international participants. Regarding their prior experience with MMGs, 50.8\% ($n=34$) had less than one year of experience, 20.9\% ($n=14$) had one to two years, 17.9\% ($n=12$) had three to five years, and 10.5\% ($n=7$) had five or more years. Details of random assignment and session structure are provided in Section 3.2 (Study Design). All participants provided their informed consent prior to the study, and the study protocol was approved by the institutional review board.

\subsection{Study Design}
\subsubsection{Overview}
We employed a between-subjects experiment to investigate the effects of clue-delivery modality. The recruited 72 participants were equally assigned to one of the two following experimental conditions: the Physical-Card condition ($n=36$, hereafter Physical-Card) and the In-App condition ($n=36$, hereafter In-App). Within each condition, the participants were further organized into six independent sessions of six players each, resulting in a total of 12 sessions. Each session involved playing our custom-designed murder mystery script, \textit{Judgment Quartet} (see Appendix A for Game Materials details), under the guidance of a custom-built social robot, POIROT, operated via a Wizard-of-Oz (WoZ) setup for movement and delivery control. Block randomization was used to maintain balanced group sizes across conditions and sessions. To ensure a highly controlled environment, each participant, regardless of condition, was provided with an iPad of identical screen size running a custom-developed companion application, which standardized the preliminary game phases of room creation, character selection, and private script reading. \rev{The experimental manipulation defined the modality through which participants received their clue materials (see Section~3.2.2).} Of all participants, 67 completed all procedures and were included in the final analyses (35 in the Physical-Card, 32 in the In-App).

\vspace{-10pt}
\subsubsection{Experimental Conditions}
All sessions began with the same app-based setup routine: participants opened the companion application, created or joined a shared room, selected one of the six predefined roles, and unlocked their private scripts on their assigned iPads (see Appendix B, Figure 1 for more details). This standardized start-up phase ensured that information asymmetry and role assignment were identical under all conditions. We then manipulated a single between-subjects factor, clue-delivery modality, with two levels: \textbf{(a) Physical-Card Condition.} After each participant unlocked their personal script, the distribution of all clue materials commenced. The WoZ operator tele-operated POIROT's mobile base to each player sequentially. Upon arriving, the robot verbally announced the delivery (e.g., "You will now receive six private clue cards") and dispensed the laminated cards directly through its clue card dispenser into the addressed player's hand as they reached forward to receive each card (see Appendix B, Figure 2 for sample cards; Appendix C, Figure 5 for dispenser technical specifications). After distributing all private clue cards, POIROT was tele-operated to move to the table center to dispense the eight public cards with a general announcement (e.g., "You will now receive eight public clue cards"). All clue delivery was completed before the players began reading the scripts. \textbf{(b) In-App Condition.} At the corresponding procedural stage (i.e., post-script unlocking), the clues were delivered digitally. After unlocking their personal script, each participant could access a "Clues" button located in the bottom-left corner of the script interface (see Appendix B, Figure 1). When pressed, this button triggered a slide-up panel displaying the player's full set of private clues. Afterwards, the eight public clues were released simultaneously to all players on the same panel (see Appendix B, Figure 3 for digital clues panel). To maintain verbal consistency with the Physical-Card condition, POIROT made identical personalized and general announcements. The app maintained real-time game state synchronization via Firebase \cite{firebase} in real time. All clue delivery was completed before the players began reading the scripts.

\rev{Except for the delivery modality, all other aspects of the interaction, including narration, robot voice, eye animations, arm gestures, and base navigation speed, were kept constant. The WoZ operator followed an identical standardized procedure in both conditions. However, quantitative analysis of session recordings indicated an inherent temporal difference: the In-App delivery was almost instantaneous ($\approx$ 3\,s), whereas the full Physical-Card delivery sequence (approaching and dispensing) required approximately 4--5 minutes on average. We treat this temporal difference as an intrinsic property of the tangible interaction modality rather than a confound.}

\vspace{-6pt}
\subsubsection{Apparatus and Environment}
\begin{figure}[htbp]
    \centering
    \includegraphics[width=0.3\textwidth, height=3cm]{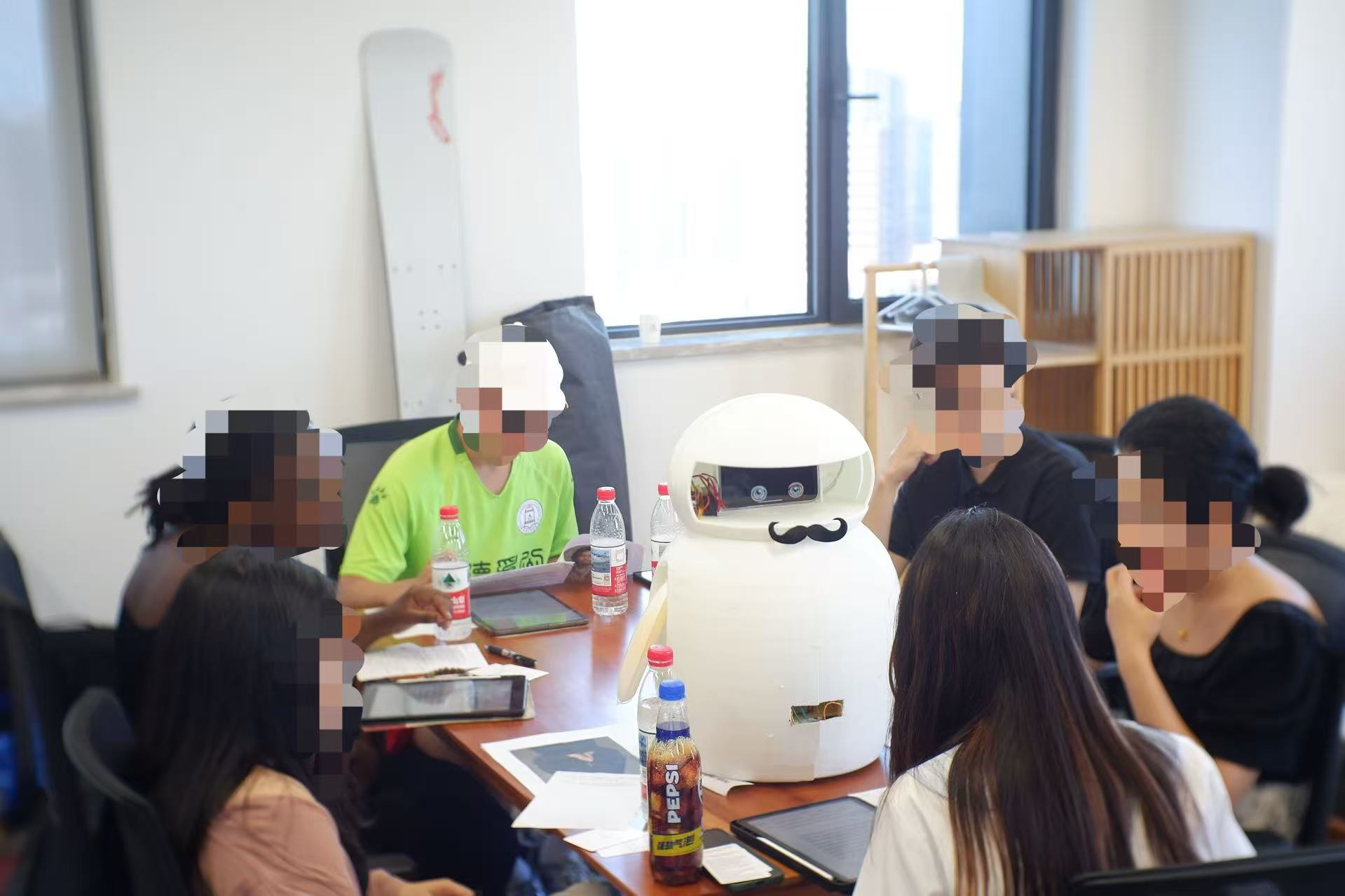}
    \Description{A picture showing the experimental setup. A group of participants sits around a rectangular dining table in a living-room-style laboratory environment. The custom-built robot, POIROT, stands on the table surface at the head of the table, facing the participants who are engaged in the game.}
    \vspace{-5pt}
    \caption{A gameplay session with POIROT.}
    \label{fig:gameplay}
\end{figure}

Each session was conducted in a quiet living-room-style laboratory environment ($\approx$7.0\,m $\times$ 8.5\,m), furnished with a central dining table ($\approx$0.75\,m $\times$ 1.5\,m) and six identical chairs, to better emulate the natural context of MMG (as illustrated in Figure~\ref{fig:gameplay}).

\noindent \textbf{\textit{POIROT} Robot.}
POIROT is a 50.1\,cm tall social robot with a 3D-printed shell, animated eyes, two 2-DOF arms, a card dispenser, and a circular base (adapted from the open-sourced FishBot platform~\cite{fishbot2025}). For voice interaction, POIROT featured an embedded audio pipeline composed of voice activity detection (VAD), automatic speech recognition (ASR), an LLM dialogue agent \cite{xiaozhi2025}, and text-to-speech (TTS) synthesis, implemented on an ESP32 audio capture module. To ensure narrative consistency, the LLM was preloaded with the complete \textit{Judgment Quartet} script and conversational guidelines, and supported natural barge-in turn-taking (see Appendix C for details).

\noindent \textbf{Companion App and Recording.}
A custom-developed companion application, running on iPads of an identical screen size (10.9-inch), was used for initial game phases and for digital clues delivery in the In-App condition (see Appendix B, Figure 3). \rev{Two 720p wide-angle webcams were positioned unobtrusively to verify the fidelity of condition implementation and quantify delivery timing. Beyond these manipulation checks, the recordings were not coded for behavioral analysis.}

\noindent \textbf{Wizard-of-Oz Control.} 
Following standard HRI protocol, a WoZ operator was seated in an adjacent room, invisible and inaudible to the participants. The operator tele-operated POIROT's mobile base using command-line inputs via ROS~2 (running on Ubuntu~22.04) to precisely control its linear and angular movement and stopping \cite{macenski2022robot}. The clue cards dispenser was activated using a mechanical push switch attached to a DuPont wire. This wire was discreetly routed to minimize visibility and participants' seating was arranged to avoid proximity to it.

\vspace{-5pt}
\subsubsection{Procedure}
Each session lasted approximately 95--100 minutes depending on the condition. Upon arrival and providing informed consent, all participants were first asked to complete a pre-study questionnaire. Afterwards, the experiment followed a four-phase structure as below:

\noindent \textbf{Introduction and Clues Distribution ($\approx$ 10--15 min).} The session began with POIROT delivering a welcome message and explaining the game rules. The participants then used the companion app to create or join in a shared room, select their character roles, and receive their private scripts. \rev{Following this, clues were delivered according to the specific condition (see Section~3.2.2), accompanied by identical verbal announcements, though the duration of this phase varied by modality.}

\noindent \textbf{Reading and Preparation ($\approx$ 25 min).} Once all participants had received their scripts and all clues, a quiet reading phase began. This allowed players to familiarize themselves with their roles and case materials in preparation for the main trial. To support note-taking if desired, each participant was also provided with a blank sheet of paper and a pencil, although their use was optional.

\noindent \textbf{Trial Deliberation ($\approx$ 50 min).} The main deliberation was moderated by POIROT acting as a judge and was structured in three stages: \rev{1) POIROT first summarized the case details and then prompted each player to introduce themselves in a fixed order. The robot used the same approach sequence (turning, gazing, and moving forward) to address each player before they spoke. 2) A second round of structured dialogue followed the same fixed order and interaction pattern, during which POIROT prompted each player to share information and their perspectives on the case. 3) After the structured rounds, POIROT announced a 30-minute free deliberation period during which participants could discuss the case freely.}

\noindent \textbf{Conclusion and Post-Study Questionnaires ($\approx$ 10 min).} POIROT revealed the canonical solution. The experiment concluded with participants finishing a post-study questionnaire.

\vspace{-8pt}
\subsection{Measures}
\rev{A combination of established, validated scales and custom-adapted questionnaires was used to measure our variables of interest. We conducted exploratory factor analysis (EFA) and calculated Cronbach's $\alpha$ for all scales to ensure construct validity and internal reliability. Unless otherwise noted, items were rated on a 7-point Likert scale (1 = "Strongly Disagree", 7 = "Strongly Agree").}

\vspace{-8pt}
\subsubsection{Pre-Study Covariates}
Participants completed two validated scales to account for participants' pre-existing traits and attitude as potential covariates.

\noindent \rev{\textbf{Big Five Inventory (BFI).}
We used the 44-item version of the Big Five Inventory \cite{john1999big} to measure participants' personality traits. Participants rated each statement on a 5-point Likert scale. Internal consistency was acceptable across all five dimensions (Cronbach's $\alpha$ range: 0.71--0.82).}

\noindent \rev{\textbf{Negative Attitudes towards Robots Scale (NARS).}
Baseline robot anxiety was measured using the original 14-item NARS \cite{nomura2006measurement}, which captures interactional distress and negative attitudes and is theoretically related to proxemic sensitivity \cite{mumm2011human} (Cronbach's $\alpha$ = 0.76).}

\vspace{-7pt}
\subsubsection{\rev{Post-Study Dependent Variables}}
Participants completed questionnaires to evaluate their experience. Full psychometric details, including factor matrices and scree plots, are provided in Appendix D.

\noindent \textbf{Narrative Immersion.}
Adapted from the Immersion Experience Questionnaire \cite{jennett2008measuring}, we measured participants' sense of immersion using a custom-developed, 15-item Narrative Immersion Scale, which captured attentional focus and emotional engagement, with several items tailored to our specific context to evaluate the robot's impact on the immersive experience. Psychometric analysis confirmed a dominant single-factor structure ($KMO = 0.84$, variance explained $= 49.6\%$) with high internal consistency (Cronbach's $\alpha$ = 0.93; see Appendix D.1).

\noindent \rev{\textbf{Social Presence.} 
To assess participants' perception of the GM as a tangible and responsive entity, we used a 5-item Social Presence Scale adapted from the Networked Minds Measure \cite{harms2004internal}. This scale focused on co-presence, attentional allocation, and behavioral interdependence. EFA results indicated good sampling adequacy ($KMO = 0.79$) and a single-factor structure explaining 52.0\% of the variance (Cronbach's $\alpha$ = 0.83; see Appendix D.2).}

\noindent \rev{\textbf{Game Experience.} 
A custom-developed core 7-item scale assessed general satisfaction and flow. It exhibited a solid one-factor structure ($KMO = 0.81$, variance explained $= 58.5\%$) and good reliability (Cronbach's $\alpha$ = 0.87). Additionally, participants in the Physical-Card condition completed a 4-item subscale probing tangibility-specific novelty and engagement (Cronbach's $\alpha$ = 0.84). This was analyzed separately (see Appendix D.3).}

\noindent \textbf{Qualitative Feedback.}
The questionnaire also included open-ended questions to gather participant reflections, which prompted participants to describe their overall feelings about the clue delivery method, provide suggestions for improvement, and recall any specific moments when the experience was impacted.

\vspace{-12pt}
\subsection{Data Analysis Strategy}
\noindent \textbf{Quantitative Analysis.}
Our analysis plan adopted a hierarchical two-step approach implemented using Python's statsmodels library.

First, to examine the baseline effect of tangibility (H1), we performed a one-way multivariate analysis of variance (MANOVA). This method was chosen as an omnibus test to detect whether the clue-delivery modality exerted a significant overall effect across the combined vector of the dependent variables (Narrative Immersion, Social Presence, and Game Experience). Considering the balanced design of our study, MANOVA provided a robust preliminary check for global differences while strictly controlling for the family-wise error rate (FWER) prior to more granular analyses.

Second, to test our primary moderation hypothesis (H2), we employed multilevel linear models (MLMs) \cite{raudenbush2002hierarchical} to account for the hierarchical structure of our data, in which participants were nested within 12 game sessions. Separate models were constructed for each dependent variable. Each model included the clue-delivery modality, NARS, and their interaction as fixed effects, with gameplay session specified as a random intercept (see Appendix E).

Significance was assessed using Wald tests ($z$-scores) derived from mixed-effects models. Upon detecting significant interactions, we decomposed them via simple slope analyses at low ($-1$ SD) and high ($+1$ SD) levels of NARS to clarify the direction of moderation. In post-hoc analyses, Bonferroni corrections were applied for multiple comparisons to ensure statistical robustness. Finally, exploratory analyses were conducted to examine potential mechanisms.

\noindent \textbf{Qualitative Analysis.}
We analyzed open-ended responses using a lightweight inductive coding process inspired by thematic analysis \cite{braun2006using}. After consolidation and anonymization, two of the authors reviewed the data to identify salient patterns regarding the robot GM, delivery modality, and subjective experiences. One researcher generated provisional codes based on participant language; a second researcher reviewed them. Differences were resolved through iterative discussion. Agreed-upon codes were grouped into higher-level themes and refined by comparing with original responses to ensure narrative accuracy. This analysis provides interpretive insight into the quantitative moderation effects rather than establishing the exhaustive prevalence of the themes.

\vspace{-6pt}
\section{Results}
We analyzed the data from 67 participants (35 in the Physical-Card condition, 32 in the In-App condition). Prior to hypothesis testing, we assessed the data for normality and homogeneity of variances. The Shapiro-Wilk test indicated that most data were normally distributed, with the exception of the Social Presence score in the Physical-Card condition ($W = 0.93$, $p = 0.03$). Brown-Forsythe tests confirmed that the assumption of homogeneity of variances was met for all dependent variables (all $p$ > 0.23).

\vspace{-6pt}
\subsection{Main Effect of Tangibility (H1)}
To test our first hypothesis (H1), we examined the overall effect of the clue-delivery modality on the three primary dependent variables combined (Narrative Immersion, Game Experience, and Social Presence) by using multivariate analysis of variance (MANOVA). The results did not show a significant overall effect of the experimental condition (Wilks' $\lambda = 0.97,\; F(3, 63) = 0.55,\; p = 0.65$). \rev{The follow-up analysis of effect sizes confirmed negligible to small differences across all outcomes (Narrative Immersion: $d=-0.24$; Social Presence: $d=-0.09$; Game Experience: $d=-0.22$).} This demonstrates no simple, direct difference between the Physical-Card and In-App conditions when considering the user experience as a whole. The lack of a main effect suggests \rev{tangibility} effects are not invariable and underscores the importance of investigating our primary moderation hypothesis (H2) to comprehend the conditions under which \rev{direct tangible interaction} matters.

\vspace{-6pt}
\subsection{Moderating Role of NARS (H2)}
To test our primary hypothesis (H2), we utilized an MLM to analyze the interaction between the experimental condition and the participants' NARS scores under the consideration of the nested data structure (see Appendix E for the full model specification). The analysis revealed that participants' pre-existing negative attitudes towards robots reliably moderated the effect of the clue-delivery modality on multiple dimensions of user experience.

\begin{figure}[t] 
    \centering
    
    \begin{subfigure}[b]{0.49\columnwidth}
        \centering        \includegraphics[width=\linewidth]{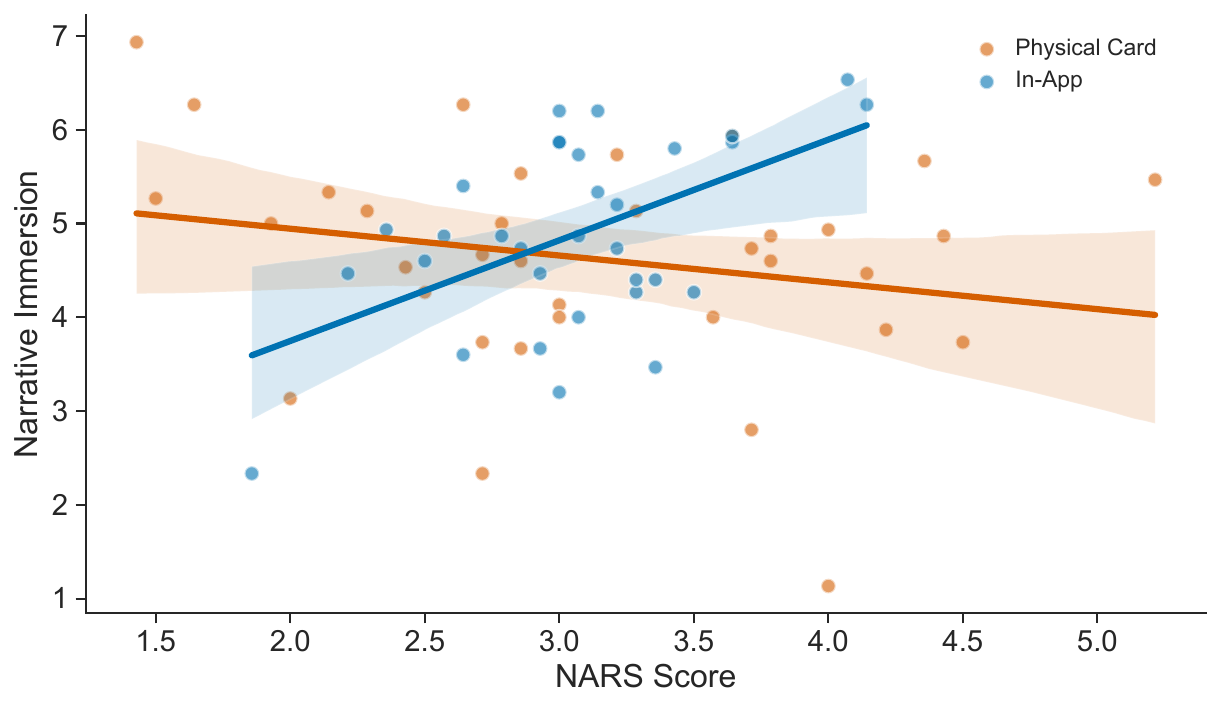}
        \caption{Narrative Immersion} 
        \Description{A line graph showing the interaction effect on Narrative Immersion. The x-axis represents NARS scores. The In-App line (orange) slopes increases significantly, indicating that as NARS increases, immersion increases sharply. The Physical-Card line (blue) is relatively flat. High NARS participants have lower immersion in the physical condition.}
        \label{fig:immersion}
    \end{subfigure}
    \hfill 
    \begin{subfigure}[b]{0.49\columnwidth}
        \centering        \includegraphics[width=\linewidth]{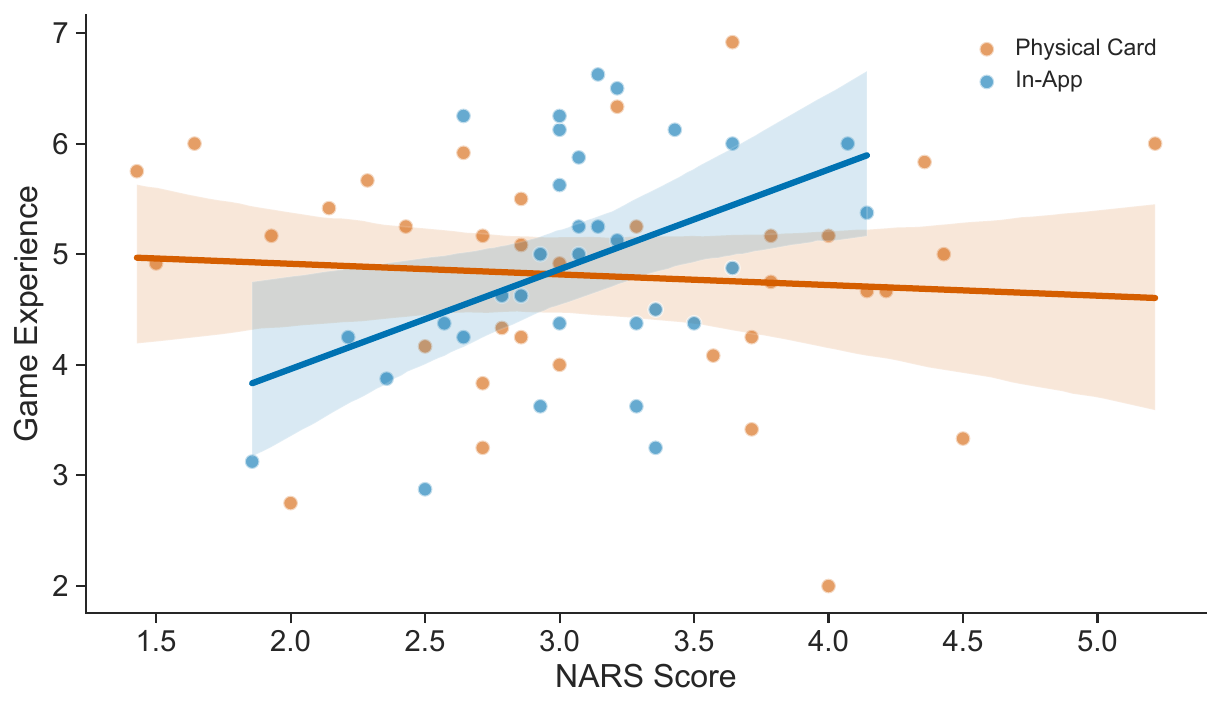}
        \caption{Game Experience}
        \Description{A line graph showing the interaction effect on Game Experience. Similar to immersion, the In-App line slopes increases as NARS increases, while the Physical-Card line remains relatively stable. The gap is widest at the high NARS end.}
        \label{fig:experience}
    \end{subfigure}
    
    \vspace{-12pt} 
    \caption{The interaction effect of experimental condition (Physical-Card vs. In-App) and participants’ Negative Attitudes towards Robots Scale (NARS) on: (a) Narrative Immersion, (b) Game Experience. Shaded areas represent 95\% confidence intervals.}
    \label{fig:interaction_effects_panel}
    
\end{figure}

\noindent \rev{\textbf{Narrative Immersion.} A significant interaction was revealed between condition and NARS ($p = 0.003$; see Table~\ref{tab:mlm_results}). To interpret this, we performed a simple slopes analysis with Bonferroni correction. For participants with high NARS scores ($+1\,SD$), the Physical-Card condition yielded significantly lower immersion compared to the In-App condition ($p_{corr} = 0.014$). Conversely, for low NARS participants ($-1\,SD$), no significant difference was observed ($p_{corr} = 0.276$). This suggests that physical delivery was detrimental only to those already wary of robots (see Table~\ref{tab:simple_slopes} and Figure~\ref{fig:immersion}).}

\noindent \rev{\textbf{Game Experience.} Similarly, a significant interaction was found for game experience ($p = 0.033$; see Table~\ref{tab:mlm_results}). The simple slopes analysis again showed that while high NARS participants trended toward lower experience in the Physical-Card condition ($p_{uncorr}$ = 0.085), this effect became non-significant after Bonferroni correction ($p_{corr}$ = 0.170). No significant difference appeared for low NARS participants ($p_{uncorr}$ = 0.254) (see Table~\ref{tab:simple_slopes} and Figure~\ref{fig:experience}).}

\noindent \textbf{Social Presence.} In contrast, the model for social presence showed neither a significant interaction effect ($p = 0.169$; see Table~\ref{tab:mlm_results}) nor a main effect of the condition.


\begin{table}[htbp]
\centering
\caption{\rev{Multilevel Models Predicting Narrative Immersion, Game Experience, and Social Presence as a Function of Condition, NARS, and Their Interaction. Random Intercepts were Specified for Game Sessions.}}
\label{tab:mlm_results}
\resizebox{\columnwidth}{!}{%
\begin{tabular}{lcccc}
\toprule
 & \textbf{b} & \textbf{SE} & \textbf{95\% CI} & \textbf{p} \\
\midrule
\multicolumn{5}{l}{\textit{Narrative Immersion}} \\
Intercept & 1.72 & 1.19 & [-0.61, 4.06] & 0.147 \\
Condition (App=0, Cards=1) & 3.65 & 1.34 & [1.01, 6.28] & 0.007 \\
NARS & 1.03 & 0.38 & [0.28, 1.78] & 0.007 \\
Condition $\times$ NARS & -1.27 & 0.43 & [-2.10, -0.44] & 0.003 \\
ICC & \multicolumn{4}{c}{0.130} \\
\midrule
\multicolumn{5}{l}{\textit{Game Experience}} \\
Intercept & 2.34 & 1.17 & [0.05, 4.64] & 0.045 \\
Condition & 2.60 & 1.32 & [0.01, 5.18] & 0.049 \\
NARS & 0.84 & 0.37 & [0.11, 1.57] & 0.025 \\
Condition $\times$ NARS & -0.88 & 0.41 & [-1.70, -0.07] & 0.033 \\
ICC & \multicolumn{4}{c}{0.183} \\
\midrule
\multicolumn{5}{l}{\textit{Social Presence}} \\
Intercept & 3.09 & 1.26 & [0.61, 5.56] & 0.014 \\
Condition & 1.76 & 1.42 & [-1.03, 4.54] & 0.216 \\
NARS & 0.57 & 0.40 & [-0.22, 1.35] & 0.156 \\
Condition $\times$ NARS & -0.61 & 0.44 & [-1.47, 0.26] & 0.169 \\
ICC & \multicolumn{4}{c}{0.292} \\
\bottomrule
\end{tabular}
}
\end{table}

\begin{table}[htbp]
\centering
\caption{\rev{Simple Slopes of Condition (Physical-Card vs. In-App) Predicting Outcomes at Low ($-1$ SD) and High ($+1$ SD) Levels of NARS. P-values are Reported as Uncorrected ($p_{unc}$) and Bonferroni-Corrected ($p_{corr}$).}}
\label{tab:simple_slopes}
\resizebox{\columnwidth}{!}{%
\begin{tabular}{llccccc}
\toprule
\textbf{Outcome} & \textbf{NARS level} & \textbf{b} & \textbf{SE} & \textbf{z} & \rev{\textbf{$\boldsymbol{p_{unc}}$}} & \rev{\textbf{$\boldsymbol{p_{corr}}$}} \\
\midrule
Narrative Immersion & Low ($-1$ SD)  & 0.66  & 0.44 & 1.49  & 0.138 & 0.275 \\
                    & High ($+1$ SD) & -1.24 & 0.46 & -2.72 & 0.007 & \textbf{0.014} \\
\midrule
Game Experience   & Low ($-1$ SD)  & 0.52  & 0.45 & 1.14  & 0.254 & 0.508 \\
                    & High ($+1$ SD) & -0.80 & 0.47 & -1.72 & 0.085 & 0.170 \\
\bottomrule
\end{tabular}%
}
\end{table}

\vspace{-8pt}
\subsection{Exploratory Analysis}
We further conducted exploratory \rev{analyses} to investigate other potential mechanisms and alternative moderators. A mediation analysis testing whether Social Presence mediated the effect of condition on Narrative Immersion was not significant; the $95\%$ confidence interval for the indirect effect included zero (ACME $95\%$ CI $[-0.47, 0.27]$). No significant moderating effects were found for any of the Big Five personality traits \cite{john1999big} or for participants' prior gaming experience (all interactions $p$ > 0.10).

\vspace{-2pt}
\subsection{Qualitative Findings: The Trade-offs of Tangible Interaction}
Based on the coding strategy from Section 3.4, we identified three themes illuminating the NARS moderation mechanism. As In-App feedback focused primarily on interface usability, we analyzed the richer experiential narratives from the Physical-Card condition regarding tangible engagement and friction. \footnote{Participant IDs (e.g., P1, P14) correspond to the 35 individuals in the Physical-Card condition. The quotations were translated and lightly edited for readability; bracketed text indicates clarifications.}.

\noindent \textbf{Theme 1: Tangibility as a Source of Novelty and Engagement.}
POIROT's most salient contribution, for many participants, was its pure novelty and the sense of occasion created by its delivery actions. The direct act of delivering clues transformed a simple game mechanic into a memorable event. One participant captured this unique quality, describing it subjectively as distinct from a digital push: "The clue distribution method is quite unique and has a very ritualistic sense" (P14). This sentiment was widely shared, with another saying that "When the robot faced me and sent me a clue card, [it] felt very ritualistic" (P21). Beyond novelty, the physical nature of the robot was considered a benefit to game fairness, with one participant appreciating that the robot was a "purer executor" and they would "not worry about the host revealing clues or being unfair" (P4). These comments suggest that for participants less sensitive to anxiety related to robots (low NARS) in particular, the direct tangible interaction functioned as a positive engagement characteristic by grounding the game in physical world and enhancing the perceived significance of the clues.

\noindent \textbf{Theme 2: Interaction Costs and Proxemic Friction.}
Alongside the positive novelty, another powerful theme centered on the interactional costs imposed by the robot's operational limitations of direct interaction. This mechanical friction provides a clear explanation for the negative experience reported by participants with high NARS. Two sources of friction came from its sluggish interaction speed, which encompassed slow physical movement and delayed verbal responses, and its artificial voice quality. The slow pace of interaction was repeatedly cited as an issue that "disrupted the flow of the game" (P8, P12). Crucially, this delay prolonged the obligatory physical engagement and forced users to maintain attention on the robot for longer than desired. Furthermore, the robot's "dramatic" electronic voice was a constant reminder of its artificiality, making one participant "feel like I'm not in the plot" (P20). From a psychological distance perspective, these moments constitute excessive proxemic demands as a result of the robot's cumbersome physical presence turning what was designed as a straightforward delivery task into a high-friction social encounter, depleting the patience and comfort of users with elevated NARS.

\noindent \textbf{Theme 3: A Desire for Social Competence.}
Finally, participants' suggestions for improvement indicated a desire for POIROT to better manage the social dynamics of tangible delivery. They looked beyond the current prototype's limitations to envision a more advanced version, a more competent social actor. Advice included enhancing its physical grace and social expressiveness, with one participant wishing "the movement can be smoother, and the logic of eyes-tracking can be more aligned with human logic" (P16). Another showed a desire for more natural, human-like social cues: "Hope it will have a mouth to speak like a human" (P11). This theme suggests that participants did not reject the idea of a robot GM; they would expect a high level of behavioral fluidity to minimize the social stress were a robot to perform direct tangible interactions. When the robot falls short of these expectations, the buffered nature of a digital interface (In-App) may become the preferred modality for its predictability and efficiency.

\vspace{-12pt}
\section{Discussion}
This study investigated how the modality of clue delivery (direct tangible vs. digitally mediated) influenced user experience in a murder mystery game. Although MANOVA showed no main effect of modality, our results revealed a crucial moderator: participants' pre-existing attitudes toward robots (NARS). This challenges the assumption that tangible interaction is universally advantageous and supports a more subtle and conditional model of HRI. By integrating quantitative and qualitative evidence, we interpret these findings and discuss their broader theoretical and practical implications.

\noindent \textbf{Theoretical Implications.}
Our findings underscore that the effect of tangible interaction in HRI is conditional rather than ubiquitous. The absence of a significant main effect is consistent with recent meta-analyses suggesting that physical presence alone does not guarantee improved engagement \cite{esterwood2025virtually, schulz2025investigating, correia2020dark}. Instead, the significant interaction effect by NARS highlights that users' attitudes toward robots act as a filter, determining whether physical proximity is experienced as an engaging feature or a social burden \cite{kabacinska2025influence, naneva2020systematic, hancock2011meta}.

\rev{Aside from the conditional effect, a primary theoretical contribution of this work is the application of a psychological distance framework to explain why users with high NARS preferred the In-App condition \cite{gittens2024remote, li2021anthropomorphism, mumm2011human}. Contrary to the viewpoint that mediated interaction is simply a "lesser form of" presence, our results suggest that digital mediation offers a distinct value: social buffering \cite{walther1996computer, walther2011theories, plomin2023virtual}. For users predisposed to robot anxiety, direct tangible delivery of clues was characterized by the robot physically navigating into their personal space and demanding handover, which likely imposed excessive proxemic demands \cite{rahmah2025modeling, obaid2016stop, koay2014social}. As our qualitative findings (Theme 2) indicate, these users perceived the robot's physical actions not as a service, but as a source of friction and forced attention \cite{rubagotti2022perceived, li2021anthropomorphism, yeh2017exploring}. Nonetheless, the In-App condition increased psychological distance, providing a safe, predictable, and low-pressure conduit for information conveyance \cite{walther1996computer, gittens2024remote, eyssel2011effects}. This validates the theoretical position that for populations with elevated robot-specific anxiety, mediated presentation can produce better experiences effectively by preserving the users' sense of control and reducing real-time coordination requirements \cite{belhassein2022addressing, shahid2024effectiveness, legler2023emotional}.}

\rev{For users holding more positive attitudes toward robots, the lack of a significant difference between conditions suggests a canceling-out effect between novelty and friction \cite{rosen2024previous, kabacinska2025influence, fischer2021tracking, smedegaard2019reframing}. As mentioned in our qualitative analysis (Theme 1), these participants appreciated the unique and real nature of the physical card delivery type, viewing the robot's tangibility as a source of engagement \cite{smedegaard2019reframing, choi2015effect, li2015benefit}. This positive novelty, however, was likely counterbalanced by the interactional costs of the physical robot (e.g., slow movement, synthetic voice), which were absent in the efficient In-App condition \cite{irfan2024human, abendschein2022novelty}. Unlike high NARS participants who viewed these costs as threatening or annoying, low NARS users tolerated them, resulting in a net experience commensurate with the smooth digital interface \cite{hoffman2014designing, kabacinska2025influence, rosen2024previous}. This implies that while tangibility offers a novelty bonus, it also incurs an interaction tax. Whether the net result is positive is heavily dependent on the users' threshold for social friction \cite{song2025no, abendschein2022novelty, smedegaard2019reframing}.}

\noindent \textbf{Practical Implications}
From a design perspective, our findings highlight key considerations for developing social robots in complex social contexts, such as collaborative games, and potentially for educational settings where interaction modality and group dynamics play a critical role \cite{lin2022benefits, gillet2020social, haripriyan2024human}. Robot game masters, exemplified by our POIROT system, should strive for adaptive mechanisms to tailor the level of proximity and mediation based upon user predispositions (e.g., pre-game NARS assessments) \cite{buracchio2025impact, gillet2020social, obaid2016stop, gillet2021robot}. In practice, designers must balance tangible novelty and social friction: Although direct physical delivery can enhance engagement through novelty and physical reality, it also incurs substantial interactional costs rooted in proxemic demands \cite{lin2022benefits, samarakoon2022review, smedegaard2019reframing}. To maximize the benefits of embodiment, designers must ensure that robotic movements are fluid enough to minimize proxemic demands and prevent the interaction from becoming socially burdensome \cite{gonccalves2024non, neto2023robot, Su2023RecentAI}. Notably, systems should support fallback mediated modes such as switching to a tablet interface for groups with higher levels of robot anxiety \cite{rosenberg2020robot, shamekhi2019multimodal}. By functioning as a social buffer, this mediated pathway can preserve user comfort without sacrificing the progression of the game, which has broader implications for HRI group settings, such as team building exercises or therapeutic games, where tangible interaction can fortify social presence, but risks alienating users who are susceptible to social pressure \cite{lin2022benefits, gillet2020social, oh2018systematic, yuliawati2024scoping}.

\vspace{-10pt}
\section{Conclusion and Future Work}
\noindent \rev{\textbf{Limitations.} First, our predominantly East Asian sample provides valuable non-WEIRD insights but limits generalizability to other demographics \cite{lim2021social, seaborn2023not, van2022weird, leichtmann2022crisis}. Future replications should examine how cultural norms influence proxemic expectations \cite{andrist2015effects, winkle2022norm, komatsu2021blaming, baxter2016characterising}. Second, the study relied on a single narrative script. Different genres or competitive stakes might alter users' tolerance for interactional costs \cite{bowman2025transformative, lopes2025closing}. For example, efficiency might be prioritized over immersion in fast-paced settings \cite{yang2021competitive, zorner2021immersive, ng2024role}. Future investigations should examine whether the social buffer of digital mediation is preferred in cooperative versus competitive tasks to an equal extent \cite{javed2024modeling, weisswange2023could}. Third, POIROT is a custom prototype; its physical movement inherently increased response latency and constituted an intrinsic temporal difference compared to the app \cite{liu2023defining, stedtler2024there}. The observed friction might thus be partially temporal. Future work should decouple tangibility from temporal duration (e.g., by introducing artificial delays) to isolate the pure effects of materiality from interaction time \cite{kappler2023optimizing, eizicovits2018robotic}. Fourth, due to the intensive session length, we prioritized NARS to specifically capture threat sensitivity and avoidance motivation, which is theoretically central to how users manage proxemic demands \cite{nomura2006measurement, nomura2007experimental, miller2021more}. Unlike general attitude scales (e.g., GAToRS) which measure broad valence, NARS provides a more direct operationalization of the anxiety-driven mechanism rendering direct tangible interaction psychologically costly for specific participants \cite{koverola2022general, nomura2007experimental}. However, we acknowledge that complementary constructs, such as RoSAS, could offer additional granularity \cite{carpinella2017robotic}. Future studies should include these measures to disentangle whether the interactional tax of tangible delivery primarily impacts competence, warmth, or trustworthiness \cite{pan2019fast, miller2021more}. Fifth, given the brief experiment session, it remains unclear whether the proxemic demands experienced by high NARS users would attenuate over time as they habituate to the presence of the robot or if the novelty bonus would fade with repeated exposure for low NARS users \cite{laban2022user, rakhymbayeva2021long, stower2021meta, ligthart2022memory}. Finally, we acknowledge a qualitative asymmetry. Physical-Card condition feedback was rich and experiential, whereas In-App's feedback was utilitarian. This imbalance prevented symmetrical thematic comparison but reflects a phenomenological reality, which indicates that screens may lack the material weight necessary to evoke the sense of occasion observed in tangible HRI while ensuring efficiency.}

\noindent \textbf{Future Directions.} Beyond addressing the limitations above, future research should further theorize interaction modality as a critical design variable in group HRI. Applying similar paradigms in healthcare or education would help test whether the social buffering effect of mediated interaction extends to other scenarios in which user anxiety is prevalent. Moreover, advancing methodological rigor through cross-laboratory replications, longitudinal designs, and open science practices will be essential to build cumulative validity in HRI \cite{strait2020three, gunes2022reproducibility, irfan2025lifelong, leichtmann2022crisis}.

\noindent \textbf{Conclusion.} This study investigated how a robot's interaction modality for delivering game content (direct tangible vs. digitally mediated) impacted user experience in a collaborative multi-party murder mystery game. Although no significant main effect of delivery modality was observed, participants' experiences were reliably moderated by NARS: for users with positive attitudes, direct tangible interaction was received neutrally or positively, whereas the same interaction undermined immersion and game experience for users with high NARS. Synthesizing our findings, this work makes three distinct contributions. Empirically, it establishes evidence from a novel controlled comparison of direct versus mediated interaction in a complex co-located context. Theoretically, we advance a psychological distance framework to demonstrate how interaction modality acts as a double-edged sword: direct tangibility provides engagement through physical reality but imposes higher proxemic demands, though digital mediation serves as a protective social buffer for anxious users. From a design standpoint, our work highlights the limits of "one-size-fits-all" systems and underscores the need for adaptive social robots that can tailor their interaction modality by switching between different channels to accommodate the diverse psychological profiles of groups.

\vspace{-12pt}

\begin{acks}
This work was supported by the Beijing Natural Science Foundation-Youth Project (Grant No. 4254082). We thank the participants and those who assisted with the study logistics. Part of the language editing was supported by Gemini 3 Pro. We also appreciate the anonymous reviewers for their constructive feedback.
\end{acks}

\balance
\bibliographystyle{ACM-Reference-Format}
\bibliography{sample-base}

\appendix

\section{Game Materials}
\label{judgment quartet}
\subsection{Plot Overview (No Spoilers)}
\label{plot overview}
Set in winter~2005 in Lubin City, the story begins with a murder during a church fund-raising event and gradually unveils the intertwined pasts of six key figures. Truth emerges only if players collaborate, challenge one another, and reconstruct the full timeline. The task focus is to co-construct the truth rather than simply identifying a murderer.


\subsection{Role Overview}
\label{role overview}
The experimental script \emph{Judgment Quartet} involves six players, each of whom plays a fixed role and is immersed in a fictional court case. Five of the six characters were part of the same university cohort, while the defendant (Fang Qi) stood apart with no prior social ties. Each character has their own identity and core motivation, as shown in Table~\ref{tab:roster}:

\begin{table}[h]
\setlength{\tabcolsep}{13pt}
\caption{Six‐Character Roster for \textit{Judgment Quartet}}
\centering\footnotesize
\label{tab:roster}
\begin{tabularx}{\linewidth}{l l X}   
\toprule
\textbf{Role} & \textbf{Identity} & \textbf{Core Motivation}\\
\midrule
Fang~Qi      & Defendant          & Maintains innocence\\
Tang~Yali    & Victim's daughter  & Justice for her father\\
Ye~Caiyang   & Prosecutor         & Legal conviction\\
Yin~Hong     & Defence atty.      & Overturn guilty verdict\\
Chu~Mihe     & Nun / Witness      & Reveal complete truth\\
Chen~Haotong & Police officer     & Support prosecution\\
\bottomrule
\end{tabularx}
\end{table}

\subsection{Clue Design and Information Matrix}
\label{clues}

\subsubsection{General Schema} 
Each player received private clue cards containing privileged memories or investigative materials. Public testimonies were released by the robot judge after private clues delivery. Sample private clues for each role are provided in Table~\ref{tab:clues map}.

\begin{table}[H]
\caption{Sample private clues provided to each role}
\label{tab:clues map}
\centering\footnotesize
\begin{tabularx}{\linewidth}{l X}
\toprule
\textbf{Role} & \textbf{Example Private Clue Excerpts}\\
\midrule
Ye~Caiyang (Prosecutor) & Vehicle CCTV logs, paint-scratch analysis, soil comparison, burial-path reconstruction\\
Tang~Yali (Victim's Daughter) & Chapel layout constraints, dead-end passages, terrain preventing escape\\
Fang~Qi (Defendant) & Fragmented memories of the parking lot\\
Chu~Mihe (Nun / Witness) & Records of prior disturbances, detailed timeline of the evening\\
Yin~Hong (Defence Lawyer) & Hidden evidence: trunk mat with victim's urine traces\\
Chen~Haotong (Police) & Cause-of-death analysis, rope-soil forensics, missing earring, suspect bruises\\
\bottomrule
\end{tabularx}
\end{table}

\subsubsection{Design Principles}
\begin{enumerate}[nosep]
  \item Mild contradictions across roles were introduced to trigger cross‐checking.
  \item Asymmetric information prevented any shared perspective.
  \item Factual evidence was balanced with emotional bias.
  \item Collaborative reasoning was promoted by piecing out truth as a whole.
\end{enumerate}

\subsubsection{Public Testimony Highlights} 
Public testimony statements included witnesses' recollections of car movements, last-seen timings, and suspicions of a third person entering the church.

\section{Setup Flow and Stimuli Illustrations}
\label{app:setup}
\begin{figure}[ht]
  \centering
  \includegraphics[width=0.22\textwidth]{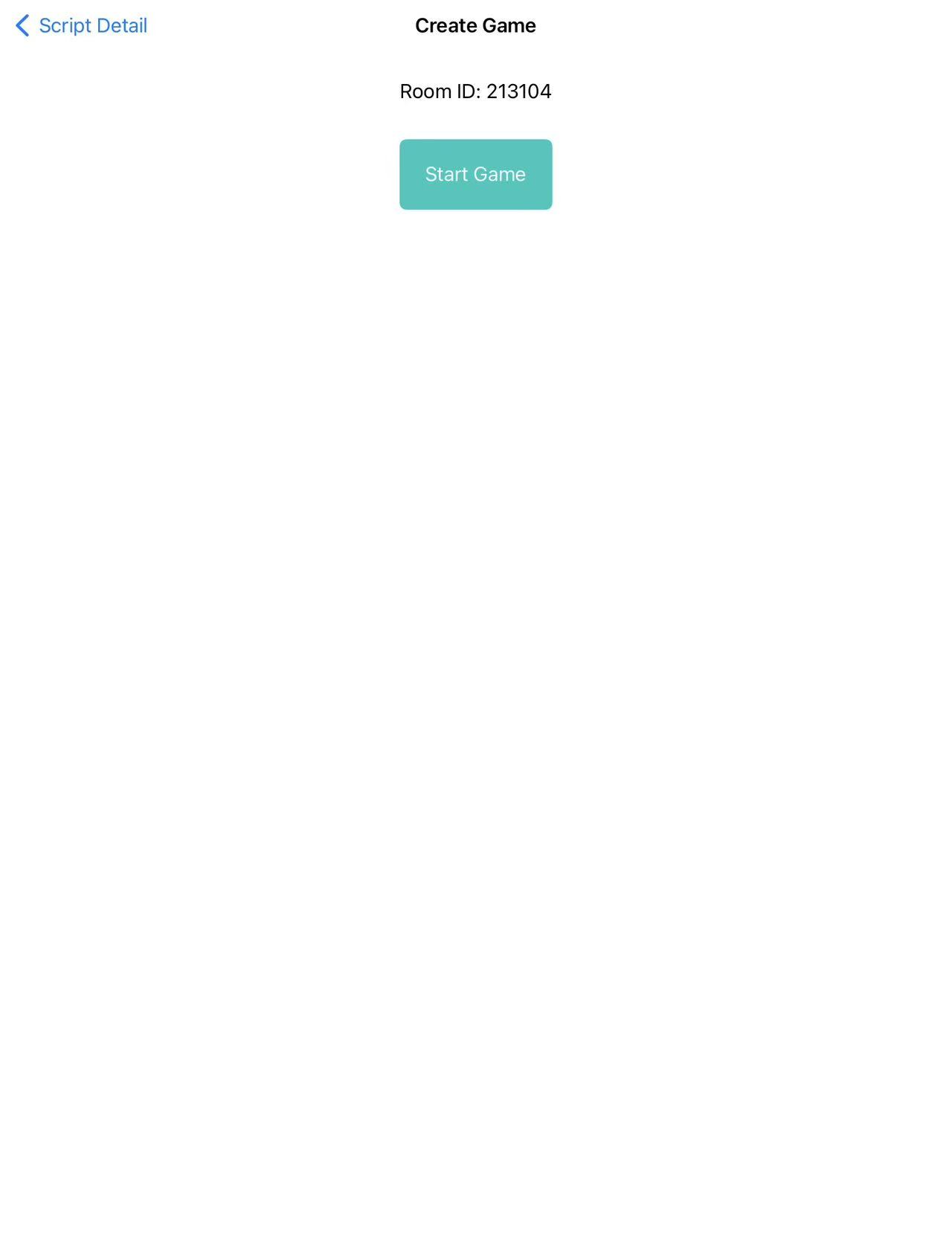}
  \includegraphics[width=0.22\textwidth]{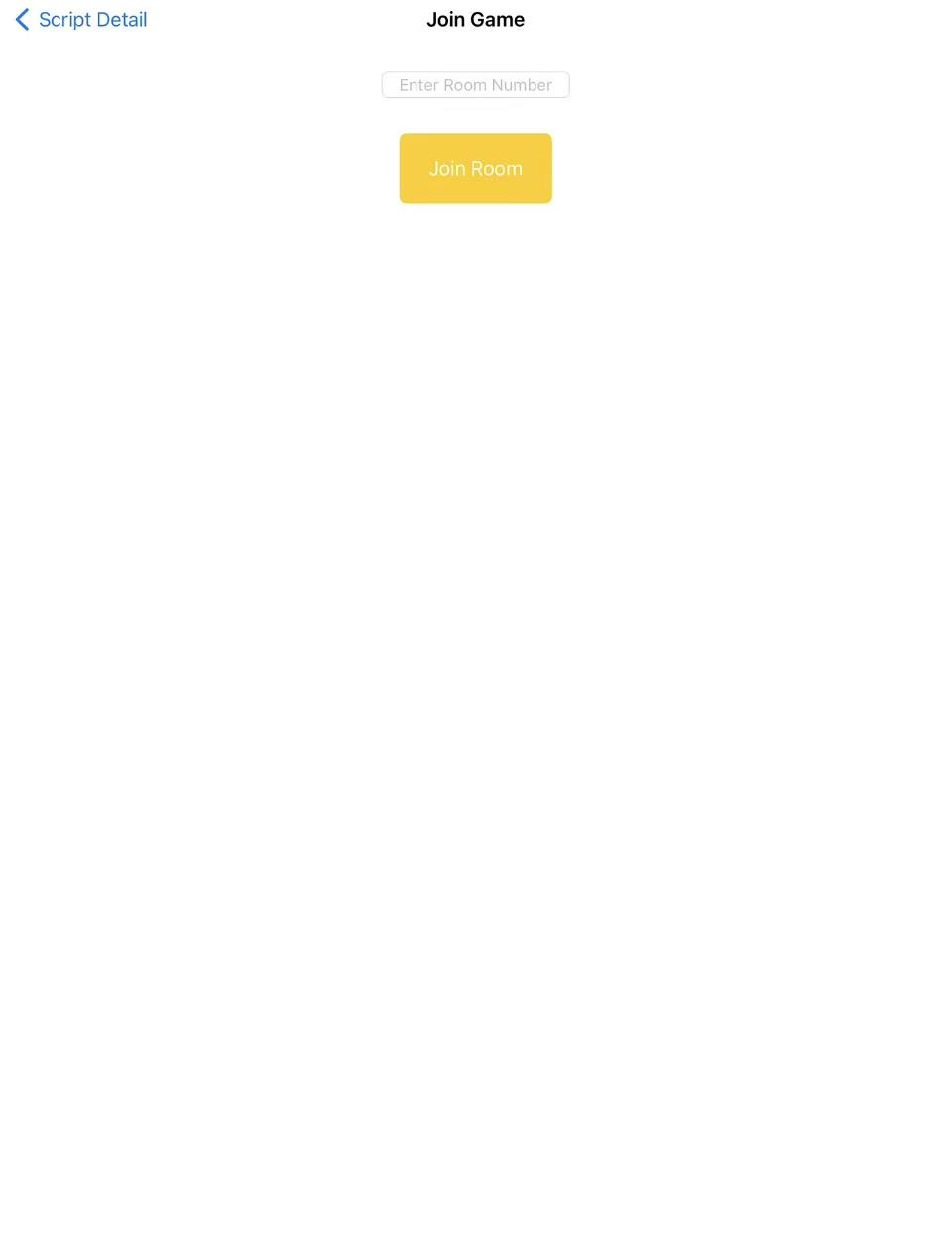}
  \includegraphics[width=0.22\textwidth]{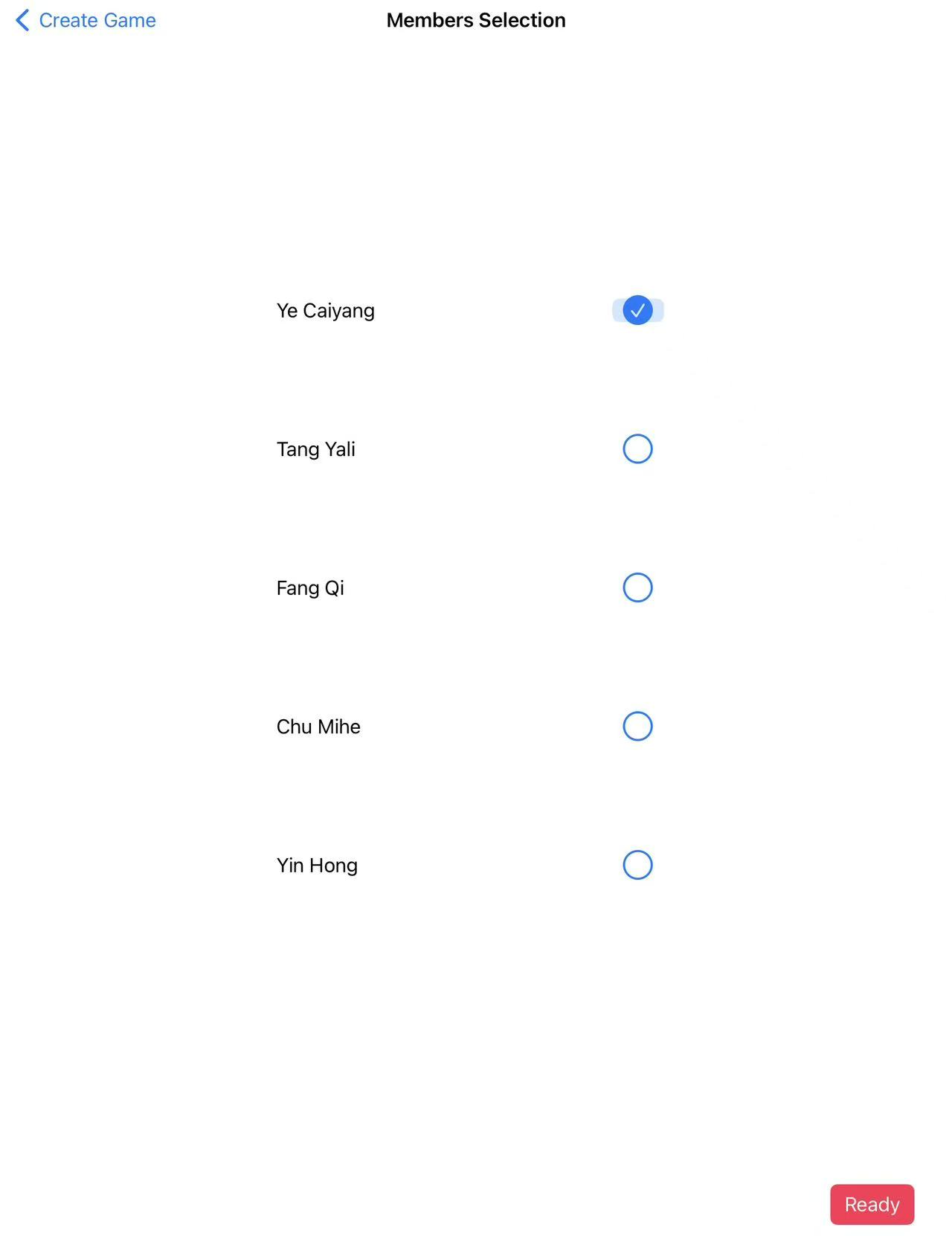}
  \includegraphics[width=0.22\textwidth]{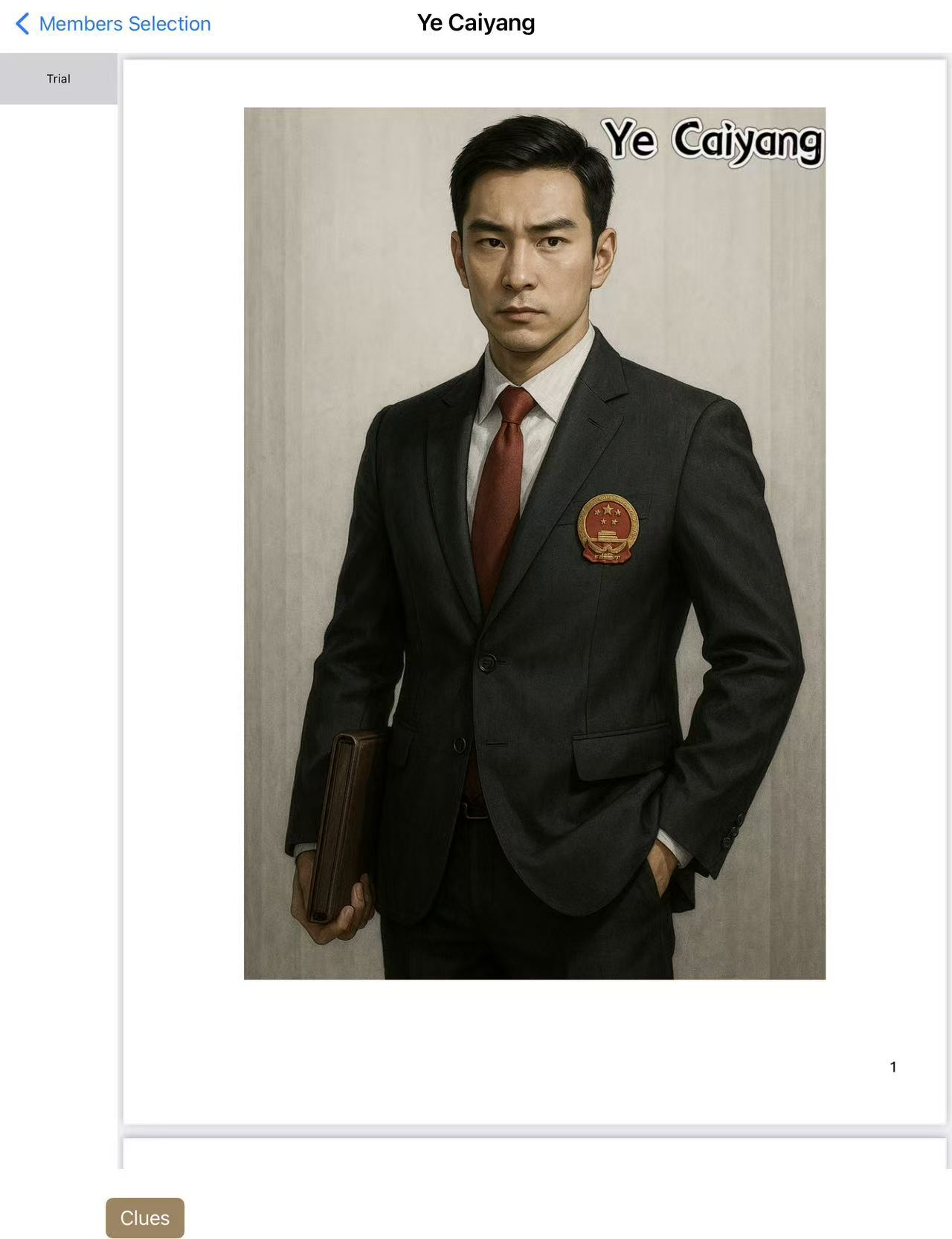}
  \Description{Four sequential screenshots of the mobile app interface. (a) shows a "Create Game" screen with a green button and a Room ID. (b) shows a "Join Game" screen with an input field. (c) displays a list of character names with radio buttons for selection. (d) shows a sample of detailed script page (featuring a photo of a male character in a suit named Ye Caiyang.)}
  \caption{Companion app setup inter
  face in sequence: (a) room creation interface, (b) room joining interface, (c) role selection interface, and (d) example of a script reading interface (see Section~3.2.3 for details). 
  \textit{Note:} In the Physical-Card condition, the "Clues" button was also present in the script reading interface but inactive, since all clues were provided via physical delivery.}
  \label{fig:app_setup}
\end{figure}

\begin{figure}[ht]
  \centering
  \includegraphics[width=0.18\textwidth]{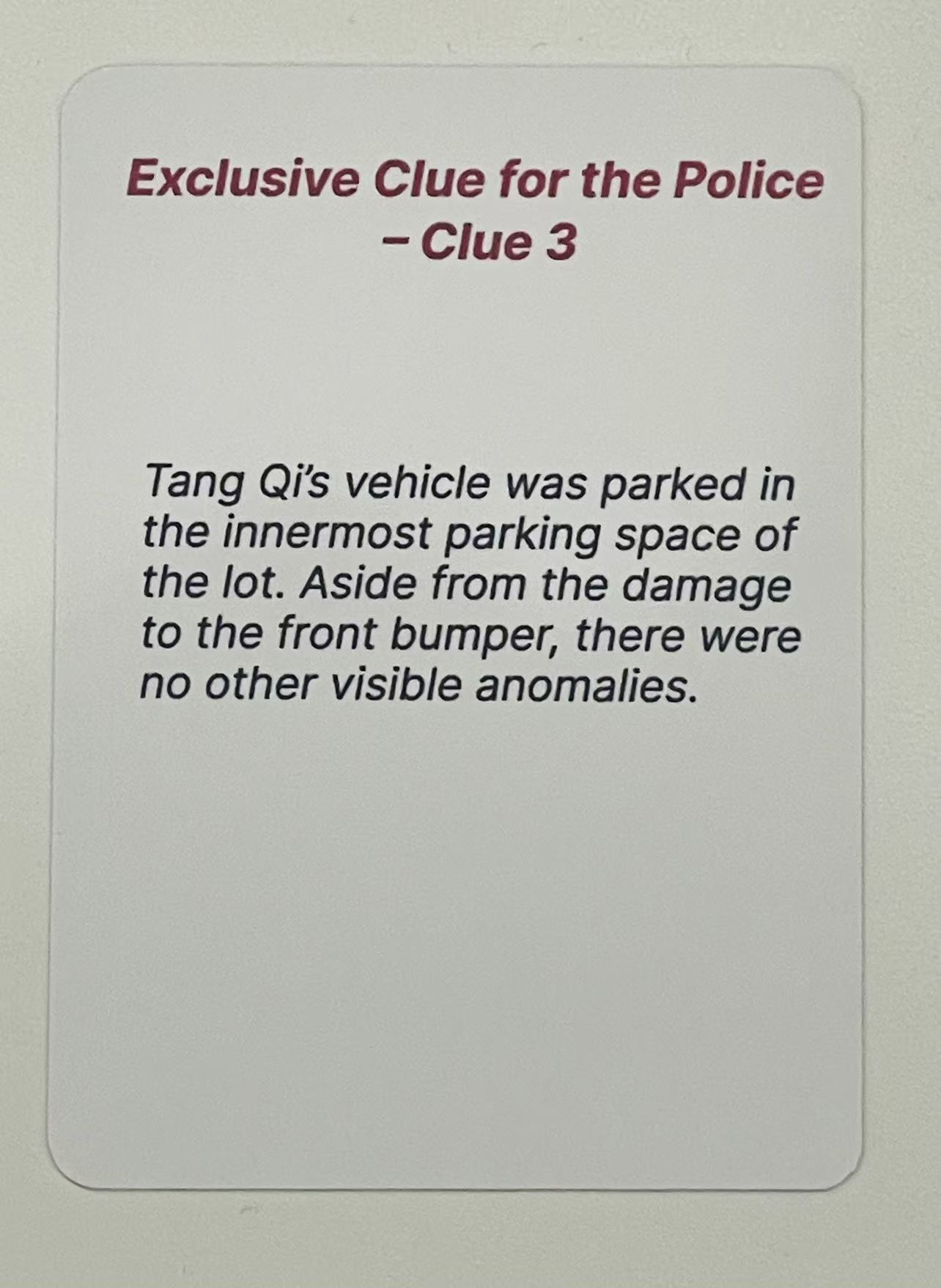}
  \includegraphics[width=0.18\textwidth]{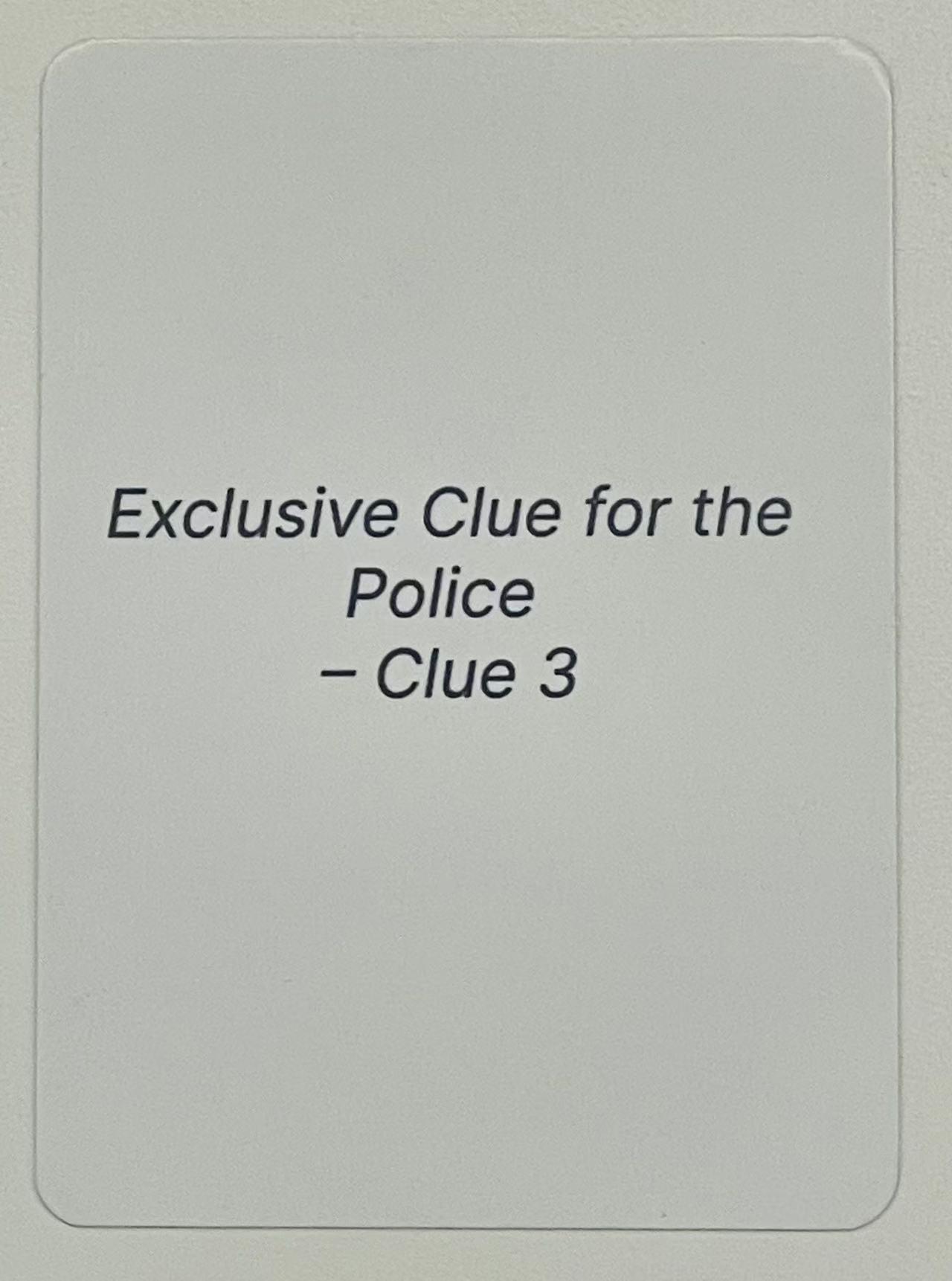}
  \Description{A picture showing the front and back of a sample laminated clue card. The left image (front) contains text labeled "Exclusive Clue for the Police -- Clue 3" followed by narrative text about a vehicle. The right image (back) shows the same label on a plain gray background without the narrative text.}
  \caption{Examples of physical clue cards (6.3\,cm $\times$ 8.8\,cm, laminated): (a) front side showing clue content, and (b) reverse side with neutral labeling. Private clues were dispensed individually to players, while public cards (eight per session) were dispensed at the table center before script reading started.}
  \label{fig:physical_clues}
\end{figure}

\begin{figure}[ht]
  \centering
  \includegraphics[width=0.18\textwidth]{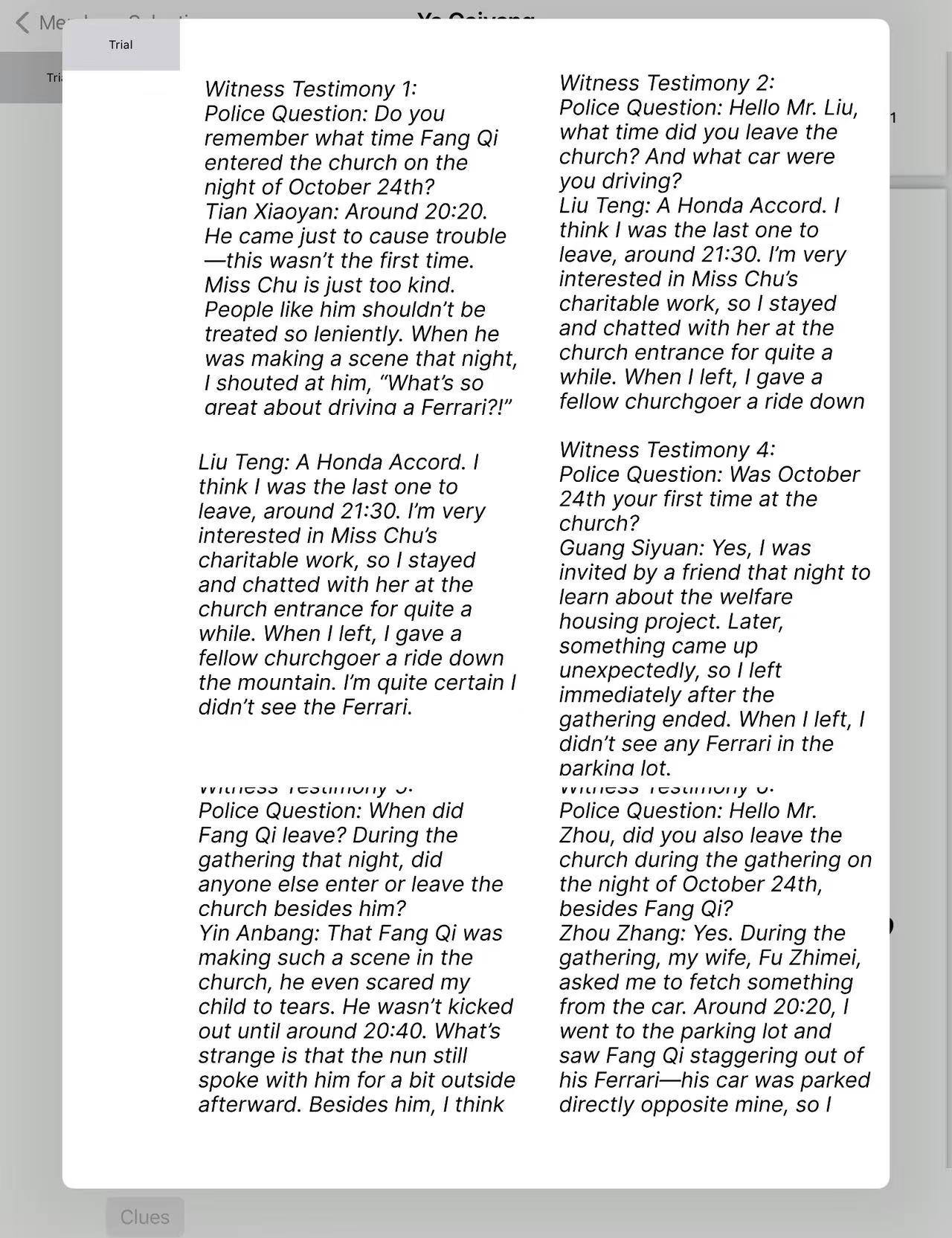}
  \includegraphics[width=0.18\textwidth]{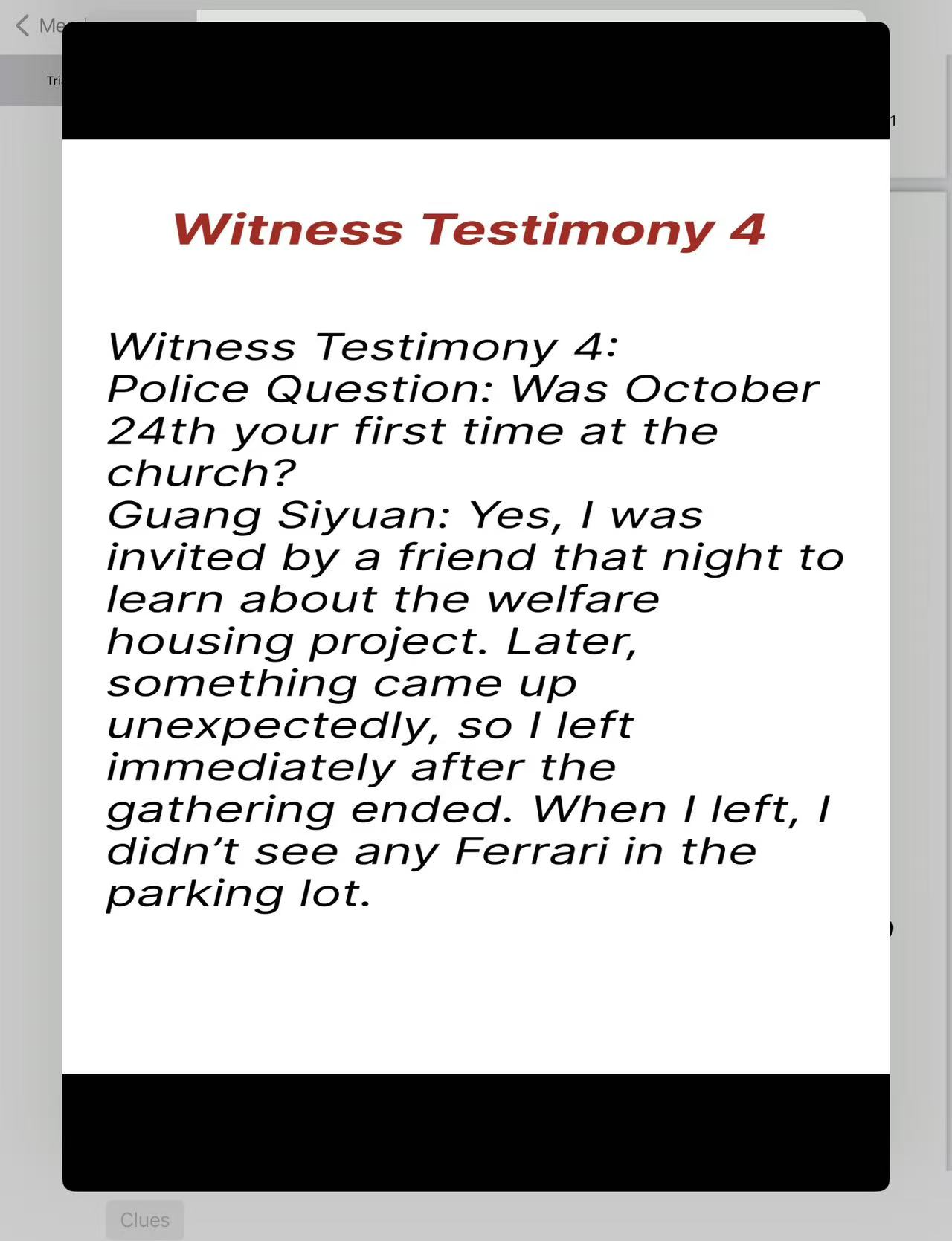}
  \Description{Two screenshots of the digital clue interfaces on a tablet. The left image shows a scrollable gallery view with two columns of text blocks representing different testimonies. The right image shows a detailed view where a single clue titled "Witness Testimony 4" is displayed in large text.}
  \caption{Digital clues panel: (a) overview of multiple digital clues, and (b) detail view of a single clue, opened by tapping the corresponding item from (a).}
  \label{fig:app_clues}
\end{figure}

This appendix illustrates visual documentation of three elements in the study:
(1) the standardized companion-app setup flow shown to all participants prior to clue delivery (Figure~\ref{fig:app_setup});
(2) examples of the physical clue cards used in the Physical-Card condition (Figure~\ref{fig:physical_clues}); and
(3) the digital clues interface used in the In-App condition (Figure~\ref{fig:app_clues}).
For clarity, all the documentation is presented in the English version; during the study, participants interacted with a personalized interface in their native language. These artifacts correspond to the procedures described in Section~3.2.2 (Experimental Conditions) and Section~3.2.4 (Procedure).

\section{Mechanical and Interaction Specifications}
\label{app:specs}
\begin{figure}[ht]
  \centering
  \includegraphics[width=0.20\textwidth]{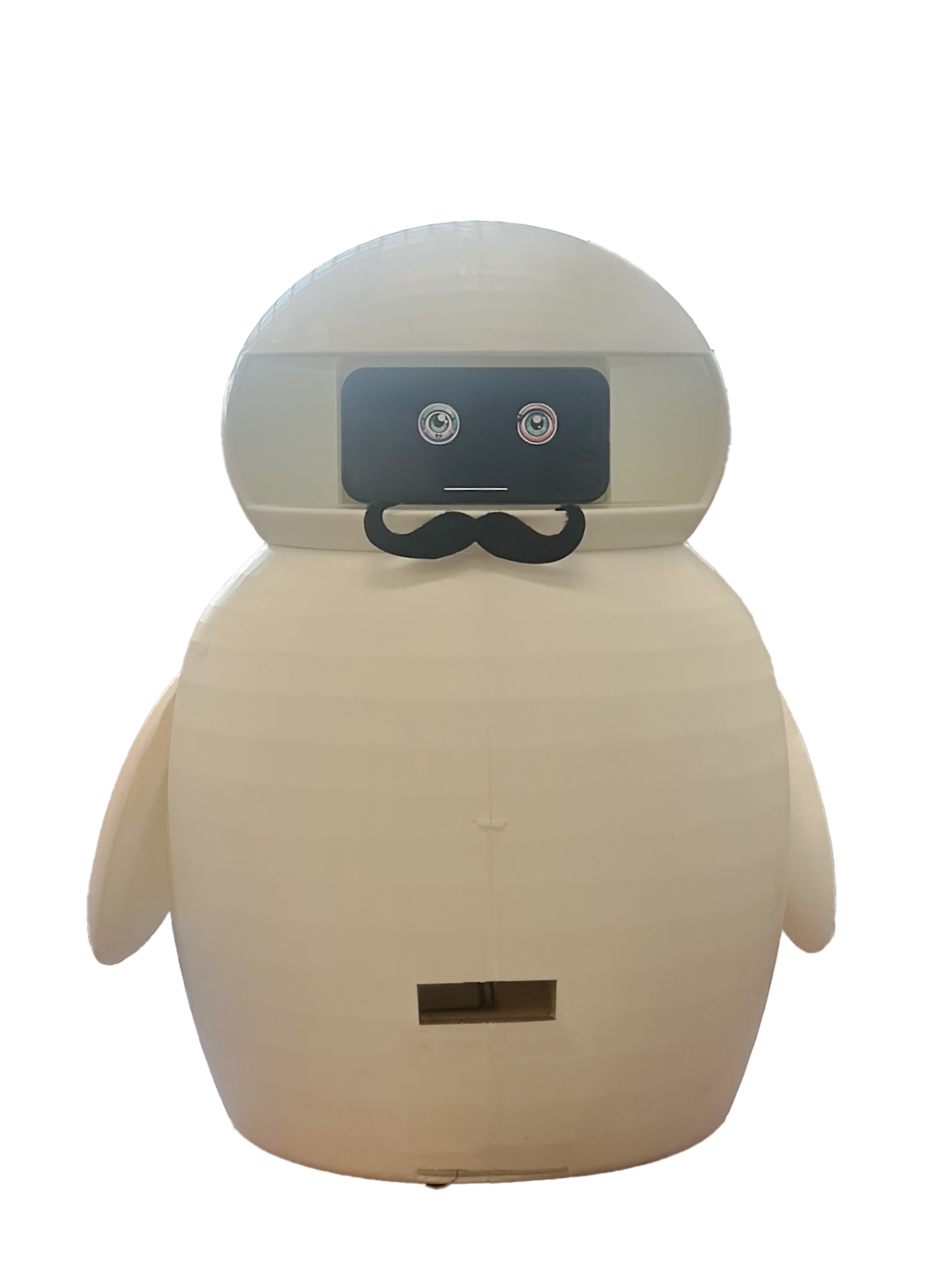}
  \Description{A front-facing picture of POIROT, which is a white, 3D-printed social robot. It has a smooth, rounded shape with small flipper arms. Its face is a black iPhone screen displaying two pixelated eyes. To evoke the style of Agatha Christie’s detective Poirot, a symbolic mustache decal was added beneath its display screen.}
  \caption{The POIROT social robot used in the study, shown in its default idle posture. To evoke the style of Agatha Christie’s detective Poirot, a symbolic mustache decal was added beneath its display screen.
}
  \label{fig:robot}
\end{figure}

This appendix provides the detailed mechanical and electrical specifications of the custom-built robot platform, POIROT (Figure~\ref{fig:robot}).

\subsection{Components Overview}
\begin{enumerate}
    \item \textit{Overall dimensions:} 50.1 cm tall; circular mobile base 34.3 cm in diameter.
    \item \textit{Shell:} 3D-printed PLA, modeled in Rhino.
    \item \textit{Locomotion:} Differential-drive base derived from the open-source FishBot platform.
    \item \textit{Arms:} Two 2-DOF arms for idle gesturing.
    \item \textit{Clue Cards dispenser:} Custom 3D-printed dispensing module embedded in the robot's abdomen. A friction roller pushed out laminated cards.
    \item \textit{Voice interaction module:} ESP32-S3 microcontroller with microphone, speaker, and amplifier supporting ASR, TTS, and dialogue management.
\end{enumerate}
Of these components, we elaborate in detail on the eye-tracking and gaze module, the clue cards dispenser, and voice interaction module, as they are most relevant to the user study. Other parts are not unfolded here since they served auxiliary or standardized functions.

\subsection{Eye-Tracking and Gaze Module}
\label{app:gaze}
We deployed the eye-tracking function on an iPhone 12 Pro device embedded in the POIROT’s head. The application used Apple’s Vision framework in Swift for real-time facial feature point detection, with a focus on the positions of both eyes. The extracted gaze data was used to dynamically adjust the robot GM’s eyes to ensure they remained aligned with participants’ faces. To enhance natural appearance, a smoothing algorithm was applied to avoid jittery eye movements, and gaze offsets were constrained within a certain range to prevent abrupt or unnatural shifts. In addition, the system simulated random blinking behaviors at intervals between three and seven seconds to conform to natural human gaze habits. Blinking animations were implemented by temporarily shrinking eye vertical height and then restoring it to the original state, producing a lifelike appearance and reducing mechanical visual artifacts. 

\subsection{Clue Cards Dispenser}
\begin{figure}[h]
  \centering
  \includegraphics[width=0.40\textwidth]{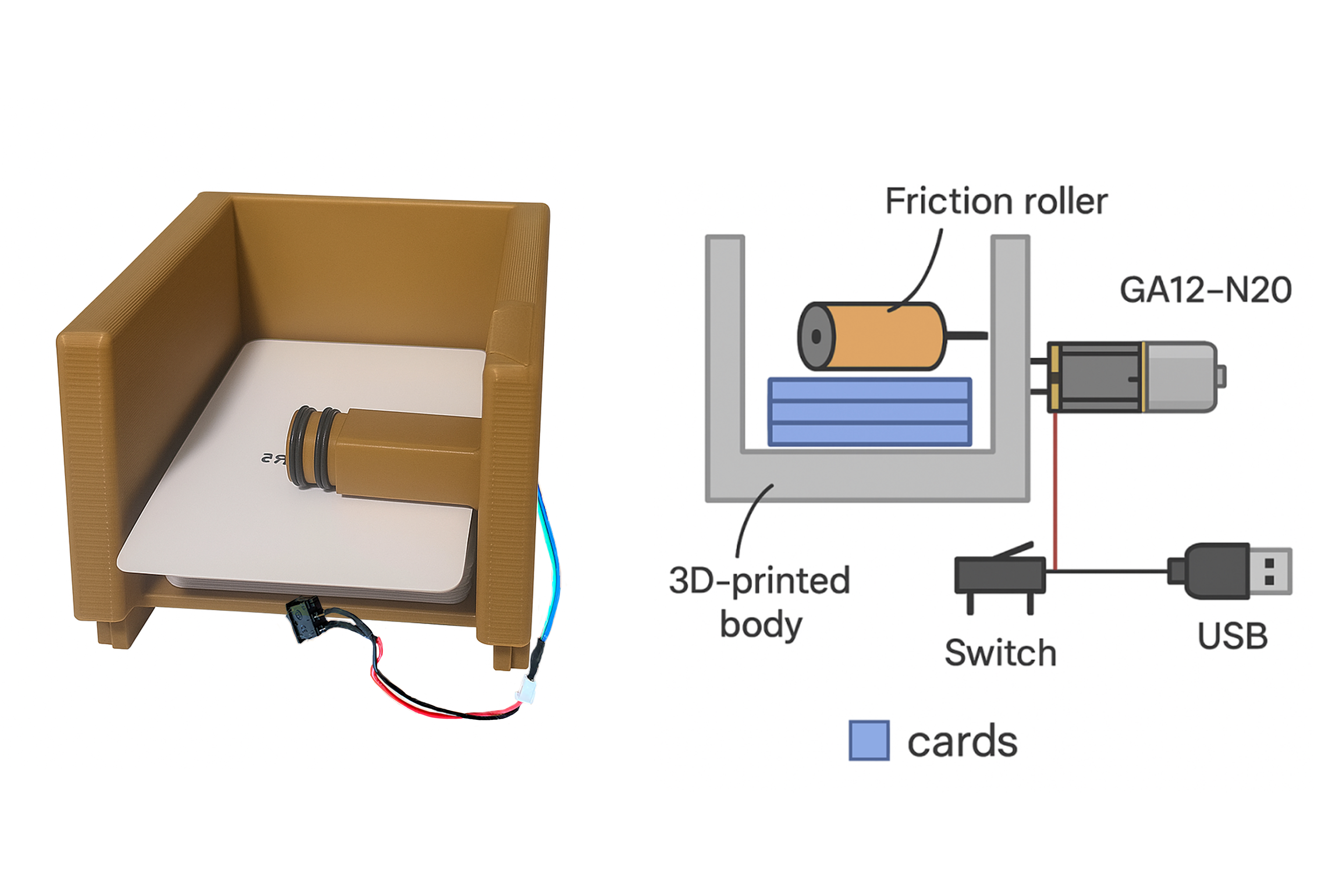}
  \Description{A composite image showing the technical design of the dispenser. The left side shows a 3D physical implementation of the clue cards dispenser containing a stack of cards and a friction roller mechanism. The right side is a schematic diagram illustrating the circuit, connecting a USB power source, a push switch, and a motor to the friction roller.}
  \caption{The custom-built clue dispensing module: (left) physical implementation with friction roller above laminated card stack, and (right) electrical diagram showing USB-powered GA12-N20 motor, friction roller, and inline SPST toggle switch. No MCU or feedback mechanism was used.}
  \label{fig:dispenser_photo}
\end{figure}

The clue cards dispenser module (Figure~\ref{fig:dispenser_photo}) was adapted for integration into the POIROT system. In this adaptation, only the upper friction-based card delivery mechanism was retained. A GA12-N20 micro gear motor (6\,V, 50\,rpm) was mounted adjacent to a 3D-printed PLA housing and mechanically coupled to a friction roller positioned above a vertically stacked array of laminated clue cards.

To enable single-card delivery per cycle, the roller was fitted with rubber O-rings, which generated sufficient frictional force during rotation to advance one card at a time. The motor received power from a standard USB-A 5\,V supply, routed through a simple SPST mechanical toggle switch that allowed manual actuation. The system design prioritized simplicity and reliability, operating without a microcontroller (MCU) or electronic feedback. Timing was manually tuned to match a consistent human-paced rhythm, with a typical dispensing duration of approximately \(1.2\,\text{s}\) per card. A complete bill of materials (BOM) is provided in Table~\ref{tab:dispenser_bom}.

\begin{table}[ht]
  \centering
  \caption{Bill of Materials for Clue Cards Dispenser Module}
  \label{tab:dispenser_bom}
  \begin{tabular}{|l|l|c|p{4.5cm}|}
    \hline
    \textbf{Component} & \textbf{Specification} & \textbf{Qty}\\
    \hline
    Motor & GA12-N20, 6V, 50rpm DC motor & 1 \\
    \hline
    Friction Roller & 3D-printed + rubber O-rings & 1  \\
    \hline
    Card Housing & PLA, 3D-printed structure & 1\\
    \hline
    Switch & SPST toggle switch & 1\\
    \hline
    Power Supply & USB-A 5V (adapter or power bank) & 1\\
    \hline
    Wiring & 2-pin cable & 1\\
    \hline
  \end{tabular}
\end{table} 

\subsection{Voice Interaction Module}
\label{voice interaction module}
The voice system was implemented on a dedicated ESP32-S3-N16R8 microcontroller with the following components:  
\begin{itemize}
    \item \textit{Audio input:} INMP441 I\textsuperscript{2}S digital microphone.  
    \item \textit{Audio output:} MAX98357 digital amplifier with a 3W speaker.  
\end{itemize}

The processing pipeline involved four stages: (1) voice activity detection (VAD), (2) streaming automatic speech recognition (ASR), (3) dialogue management and decision-making via an open-sourced LLM-based agent, \textit{Xiaozhi}, and (4) text-to-speech (TTS) synthesis. For consistency, the LLM was pre-loaded with the full script of \textit{Judgment Quartet}, including character backgrounds and conversational guidelines. A barge-in mechanism supported natural turn-taking, enabling participants to interrupt the robot's speech mid-utterance.

\subsection{Notes}
In the actual study, participants only interacted with POIROT through its observable behaviors (gaze, speech, movement, and card delivery). Further design and modeling diagrams are omitted here, as they are not directly related to the interaction study. To support transparency and reproducibility, the underlying CAD models, wiring diagrams, and control code will be released in a public repository upon acceptance of this paper.


\section{Questionnaires}
\label{appendix:questionnaires}

\subsection{Narrative Immersion Scale}
\label{survey:immersion}
\noindent
Instructions: Please indicate the extent to which you agree or disagree with each statement based on your experience in the game. (1 = Strongly disagree, 7 = Strongly agree)

\begin{enumerate}
    \item I was rarely distracted by outside factors during the game.
    \item When the robot GM asked questions, I was immediately able to engage and respond.
    \item I felt a strong emotional connection with the character I played.
    \item As the story developed, I could feel the emotional changes of the character.
    \item The robot GM’s questions and feedback deepened my emotional experience.
    \item I lost track of time during the game.
    \item After the game, it felt like time passed faster than expected.
    \item I always felt that the game time was too short and wished it could continue.
    \item During the game, I almost forgot about the real world around me.
    \item I felt as if I was completely inside the world of the character.
    \item The presence of the robot GM helped immerse me deeper into the fictional story context.
    \item I felt that my choices or decisions could influence the storyline.
    \item The way the robot GM asked questions made me feel that my role was important to the plot.
    \item I felt the robot GM’s responses were genuine rather than purely pre-programmed.
    \item Overall, I felt a strong sense of narrative immersion throughout the game.
\end{enumerate}

\rev{The Narrative Immersion Scale (15 items) demonstrated high internal consistency (Cronbach's $\alpha$ = 0.93). EFA exhibited the suitability of the data ($KMO$ = 0.84, Bartlett's Test $p$ < 0.001). A dominant single factor emerged (Eigenvalue = 7.88), explaining 49.6\% of the variance. All items showed strong factor loadings (>0.40).}

\begin{figure}[h!]
    \centering
    \begin{minipage}{0.48\textwidth}
        \centering
        \small
        \begin{tabular}{lc}
            \toprule
            \textbf{Item ID} & \textbf{Factor Loading} \\
            \midrule
            Item 1 & 0.419 \\
            Item 2 & 0.630 \\ 
            Item 3 & 0.704 \\
            Item 4 & 0.737 \\
            Item 5 & 0.847 \\
            Item 6 & 0.680 \\
            Item 7 & 0.479 \\
            Item 8 & 0.511 \\
            Item 9 & 0.734 \\
            Item 10 & 0.853 \\
            Item 11 & 0.813 \\
            Item 12 & 0.651 \\
            Item 13 & 0.792 \\
            Item 14 & 0.622 \\
            Item 15 & 0.891 \\
            \bottomrule
        \end{tabular}
        \captionof{table}{\rev{Pattern Matrix for Narrative Immersion.}}
        \label{tab:loadings_immersion}
    \end{minipage}
    \hfill
    \begin{minipage}{0.48\textwidth}
        \centering
        \includegraphics[width=\linewidth]{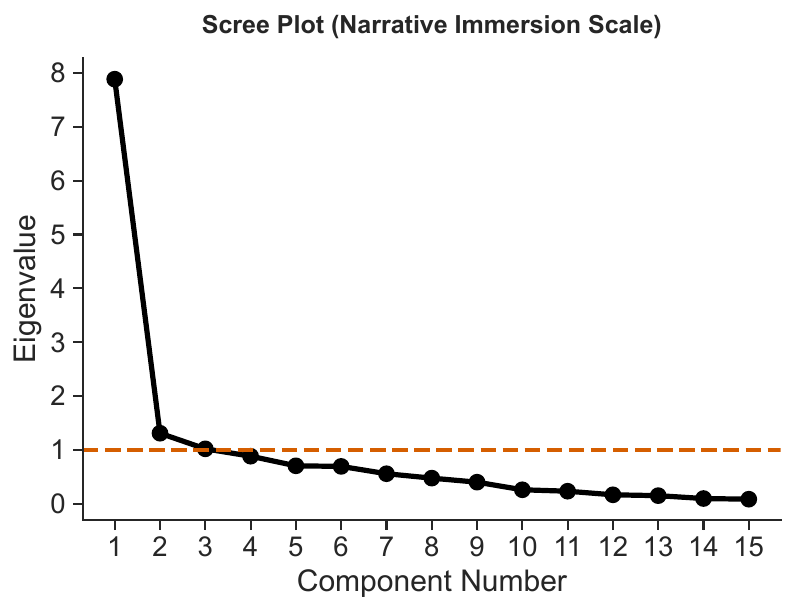}
        \Description{A scree plot showing eigenvalues for 15 components. The first point is very high at approximately 7.9, while the second point drops sharply to near 1.3, and subsequent points flatten out below 1.0, indicating a single dominant factor.}
        \caption{\rev{Scree Plot for Narrative Immersion.}}
        \label{fig:scree_immersion}
    \end{minipage}
\end{figure}

\subsection{Social Presence Scale}
\label{survey:presence}
\noindent
Instructions: Please indicate the extent to which you agree or disagree with each statement based on your experience in the game. (1 = Strongly disagree, 7 = Strongly agree)

\begin{enumerate}
    \item During the game, I felt that the GM was 'present in the same space' with me and paying attention to me.
    \item My actions or responses influenced how the GM paid attention to or responded to me afterward.
    \item I viewed the GM not just as an information provider, but as an entity with an identity and role.
    \item The GM’s reactions during the game made me feel consistently noticed rather than ignored.
    \item I was willing to accept the GM as a key host of this game and would even look forward to playing with them again.
\end{enumerate}

\rev{The Social Presence Scale (5 items) showed good reliability (Cronbach's $\alpha$ = 0.83). Data suitability was confirmed ($KMO$ = 0.79, Bartlett's Test $p$ < 0.001). Analysis revealed a clear one-factor structure explaining 52.0\% of the variance.}

\begin{figure}[h!]
    \centering
    \begin{minipage}{0.48\textwidth}
        \centering
        \small
        \begin{tabular}{lc}
            \toprule
            \textbf{Item ID} & \textbf{Factor Loading} \\
            \midrule
            Item 1 & 0.821 \\
            Item 2 & 0.723 \\
            Item 3 & 0.712 \\
            Item 4 & 0.827 \\
            Item 5 & 0.458 \\
            \bottomrule
        \end{tabular}
        \captionof{table}{\rev{Pattern Matrix for Social Presence.}}
        \label{tab:loadings_social}
    \end{minipage}
    \hfill
    \begin{minipage}{0.48\textwidth}
        \centering
        \includegraphics[width=\linewidth]{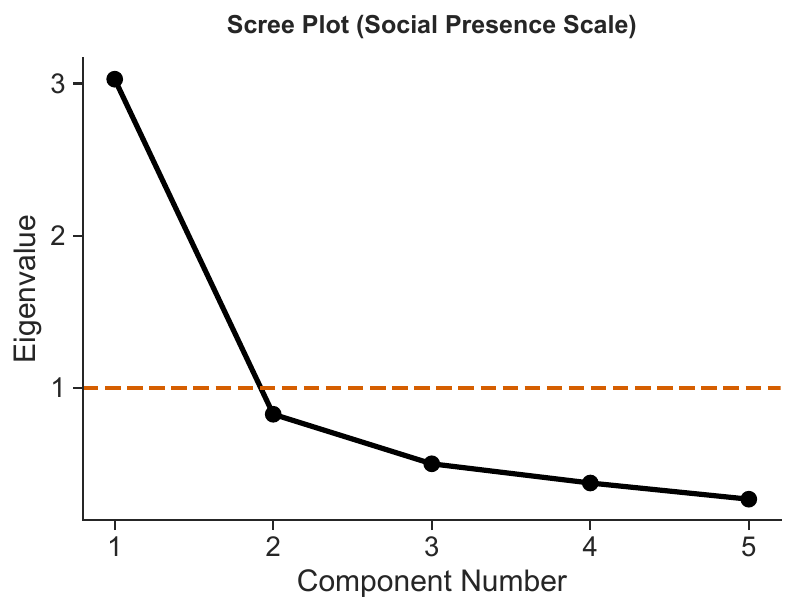}
        \Description{A scree plot for 5 components. The first component has an eigenvalue of roughly 3.0. The second component drops below the red threshold line of 1.0 (to $\approx$ 0.8), confirming a single-factor structure.}
        \caption{\rev{Scree Plot for Social Presence.}}
        \label{fig:scree_social}
    \end{minipage}
\end{figure}

\subsection{Game Experience Scale}
\label{survey:mmg}
\noindent
Instructions: Please answer honestly. There are no right or wrong answers. 

\subsubsection{Game Experience and Robot Interaction Items  (1 = Strongly disagree, 7 = Strongly agree)} 
\noindent
Items 1–7 were administered to all participants. Items 8–11 were only administered to those in the Physical-Card condition, as they specifically concerned the physical clues delivery with the robot.
\begin{enumerate}
    \item I felt deeply immersed in the storyline during the game.
    \item The clues conveyed a strong sense of presence in the story.
    \item The delivery of clues felt natural and smooth, without interrupting my gameplay.
    \item The way clues were delivered enhanced my role immersion.
    \item I found it easier to remember the content of the clues I received.
    \item I was satisfied with the efficiency of the clue delivery method.
    \item The process of receiving clues did not make me feel distracted or disconnected.
    \item I felt that the robot had a certain level of character or personality.
    \item When the robot approached and handed me a clue, I experienced a special sense of ritual.
    \item The way the robot provides information makes me feel that the information is reliable.
    \item I would like to see more games using physical robots as clue deliverers or hosts in the future.
\end{enumerate}

\rev{The Core Game Experience Scale (7 items) demonstrated high internal consistency (Cronbach's $\alpha$ = 0.87). The data exhbited good suitability for factor analysis ($KMO$ = 0.81, Bartlett's Test $p$ < 0.001), yielding a single-factor structure that explained 58.5\% of the variance. Note that the 4 additional items specific to the Physical-Card condition are excluded from this factor analysis. Regarding the Tangibility Subscale (Items 8--11), which was administered exclusively to the Physical-Card condition, these items also demonstrated high internal consistency (Cronbach's $\alpha$ = 0.84). They were excluded from the main factor analysis as they were not completed by the full sample.}

\begin{figure}[h!]
    \centering
    \begin{minipage}{0.48\textwidth}
        \centering
        \small
        \begin{tabular}{lc}
            \toprule
            \textbf{Item ID} & \textbf{Factor Loading} \\
            \midrule
            Item 1 & 0.751 \\
            Item 2 & 0.825 \\
            Item 3 & 0.888 \\
            Item 4 & 0.904 \\
            Item 5 & 0.547 \\
            Item 6 & 0.696 \\
            Item 7 & 0.681 \\
            \bottomrule
        \end{tabular}
        \captionof{table}{\rev{Pattern Matrix for Game Experience (Core).}}
        \label{tab:loadings_game}
    \end{minipage}
    \hfill
    \begin{minipage}{0.48\textwidth}
        \centering
        \includegraphics[width=\linewidth]{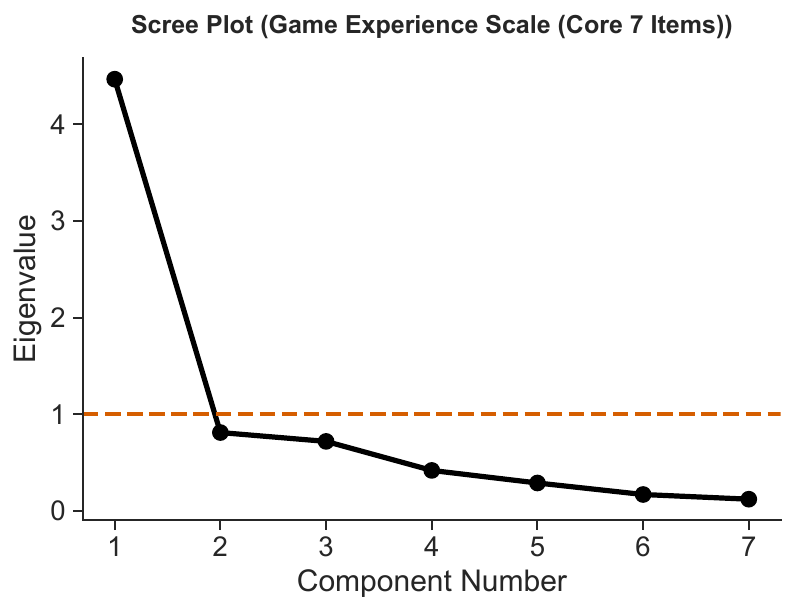}
        \Description{A scree plot for 7 components. The first component starts high at approximately 4.7. The second component drops significantly below 1.0 (to $\approx$ 0.8), followed by a flat tail, indicating one main factor.}
        \caption{\rev{Scree Plot for Game Experience.}}
        \label{fig:scree_game}
    \end{minipage}
\end{figure}

\subsubsection{General Feedback (Open-ended)}
\begin{enumerate}
    \item Please briefly describe your overall impression of the clue delivery method used in this game.
    \item Have you ever been confused by the form or timing of clue distribution? Does this affect your reasoning rhythm or immersion?
    \item In this game, is there a clue distribution that impressed you? Please describe the situation and your reaction.
    \item What suggestions do you have or hope to improve on the form, rhythm or expression of clue distribution?
\end{enumerate}

\section{Multilevel Linear Model Specification}
To analyze each participant $i$ nested within gameplay session $j$, we employed a multilevel linear model to predict outcome variables (e.g., Narrative Immersion) as follows:
\label{mlm}
\begin{align}
Y_{ij} = \ & \beta_0 + \beta_1 \cdot \text{Session}_{ij} + \beta_2 \cdot \text{Moderator}_{ij} \nonumber \\
          & + \beta_3 \cdot (\text{Session}_{ij} \times \text{Moderator}_{ij}) + u_j + \epsilon_{ij}
\end{align}

\noindent
Where:
\begin{itemize}
    \item $Y_{ij}$: outcome score (e.g., Narrative Immersion) for participant $i$ in session $j$
    \item $\text{Session}_{ij}$: experimental condition (0 = In-App, 1 = Physical-Cards)
    \item $\text{Moderator}_{ij}$: moderator variable (e.g., NARS or Big Five personality traits)
    \item $\text{Session}_{ij} \times \text{Moderator}_{ij}$: interaction term for testing moderation effects
    \item $u_j \sim \mathcal{N}(0, \sigma^2_u)$: random intercept for each gameplay session
    \item $\epsilon_{ij} \sim \mathcal{N}(0, \sigma^2)$: individual-level residual
\end{itemize}

\end{document}